\documentclass[sigconf]{acmart}
\usepackage{multirow}
\usepackage{algorithm}
\usepackage{algorithmic}
\usepackage{booktabs}
\usepackage{float}
\pdfoutput=1
\renewcommand\footnotetextcopyrightpermission[1]{} % removes footnote with conference information in first column
\AtBeginDocument{%
  \providecommand\BibTeX{{%
    \normalfont B\kern-0.5em{\scshape i\kern-0.25em b}\kern-0.8em\TeX}}}

\setcopyright{acmcopyright}
\copyrightyear{2023}
\acmYear{2023}
\acmDOI{10.1145/1122445.1122456}

\begin{document}

%\title{Trustworthy Open-world Learning Framework Exploration: From a Universal Exxiwperience Perspective}

\title{Bridging Trustworthiness and Open-World Learning: \\An Exploratory Neural Approach for Enhancing \\ Interpretability, Generalization, and Robustness}
%Trustworthy Learning Framework Exploration: Enhancing Trust in Learning Experiences
%Trustworthy Neural Networks with Interpretability, Generalization and Robustness: From a Universal Perspective

\author{Shide Du}
\email{dushidems@gmail.com}
\affiliation{%
  \institution{Fuzhou University}
  \city{Fuzhou}
%\state{Fujian}
  \country{China}
}

\author{Zihan Fang}
\email{fzihan11@163.com}
\affiliation{%
  \institution{Fuzhou University}
  \city{Fuzhou}
%\state{Fujian}
  \country{China}
}

\author{Shiyang Lan}
\email{llanshiyang@163.com}
\affiliation{%
  \institution{Fuzhou University}
  \city{Fuzhou}
%\state{Fujian}
  \country{China}
}

\author{Yanchao Tan}
\email{yctan@fzu.edu.cn}
\affiliation{%
  \institution{Fuzhou University}
  \city{Fuzhou}
%\state{Fujian}
  \country{China}
}

\author{Manuel Günther}
\email{siebenkopf@googlemail.com}
\affiliation{%
  \institution{University of Zurich}
  \city{Zurich}
%\state{Zurich}
  \country{Switzerland}
}

\author{Shiping Wang}
\email{shipingwangphd@163.com}
\affiliation{%
  \institution{Fuzhou University}
  \city{Fuzhou}
%\state{Fujian}
  \country{China}
}

\author{Wenzhong Guo}
\email{fzugwz@163.com}
\affiliation{%
  \institution{Fuzhou University}
  \city{Fuzhou}
%\state{Fujian}
  \country{China}
}
\authornote{Corresponding author.}

\renewcommand{\shortauthors}{Shide Du et al.}

\begin{abstract}
As researchers strive to narrow the gap between machine intelligence and human through the development of artificial intelligence technologies, it is imperative that we recognize the critical importance of trustworthiness in open-world, which has become ubiquitous in all aspects of daily life for everyone.
However, several challenges may create a crisis of trust in current artificial intelligence systems that need to be bridged: 1) Insufficient explanation of predictive results; 2) Inadequate generalization for learning models; 3) Poor adaptability to uncertain environments.
Consequently, we explore a neural program to bridge trustworthiness and open-world learning, extending from single-modal to multi-modal scenarios for readers.
%1) With the aid of optimization problem-derived technologies, we formalize the objective functions with specific physical meanings to customize trustworthy networks in a transparent manner and heighten design-level interpretability; 
%2) By environmental well-being task-interfaces with flexible learning regularizers, the proposed frameworks strengthen the perception capabilities of models to increase the generalization of trustworthy learning;
%3) Open-world-designed recognition losses and agent mechanisms endow the abilities on the models in handling unknown data to promote the robustness of trustworthy learning.
1) To enhance design-level interpretability, we first customize trustworthy networks with specific physical meanings; 
2) We then design environmental well-being task-interfaces via flexible learning regularizers for improving the generalization of trustworthy learning; 
3) We propose to increase the robustness of trustworthy learning by integrating open-world recognition losses with agent mechanisms.
Eventually, we enhance various trustworthy properties through the establishment of design-level explainability, environmental well-being task-interfaces and open-world recognition programs. 
These designed open-world protocols are applicable across a wide range of surroundings, under open-world multimedia recognition scenarios with significant performance improvements observed.
%As a result, this work provides valuable insights for interested developers and accelerates research on the critical issues of trustworthiness, exploration and solution.
\end{abstract}


%\ccsdesc[500]{Computing methodologies~Unsupervised learning}
%\ccsdesc[500]{Computing methodologies~Learning latent representations}


\maketitle

\section{Introduction}\label{sec:int}

\begin{figure}[t]
  \centering
  \includegraphics[width=0.48\textwidth]{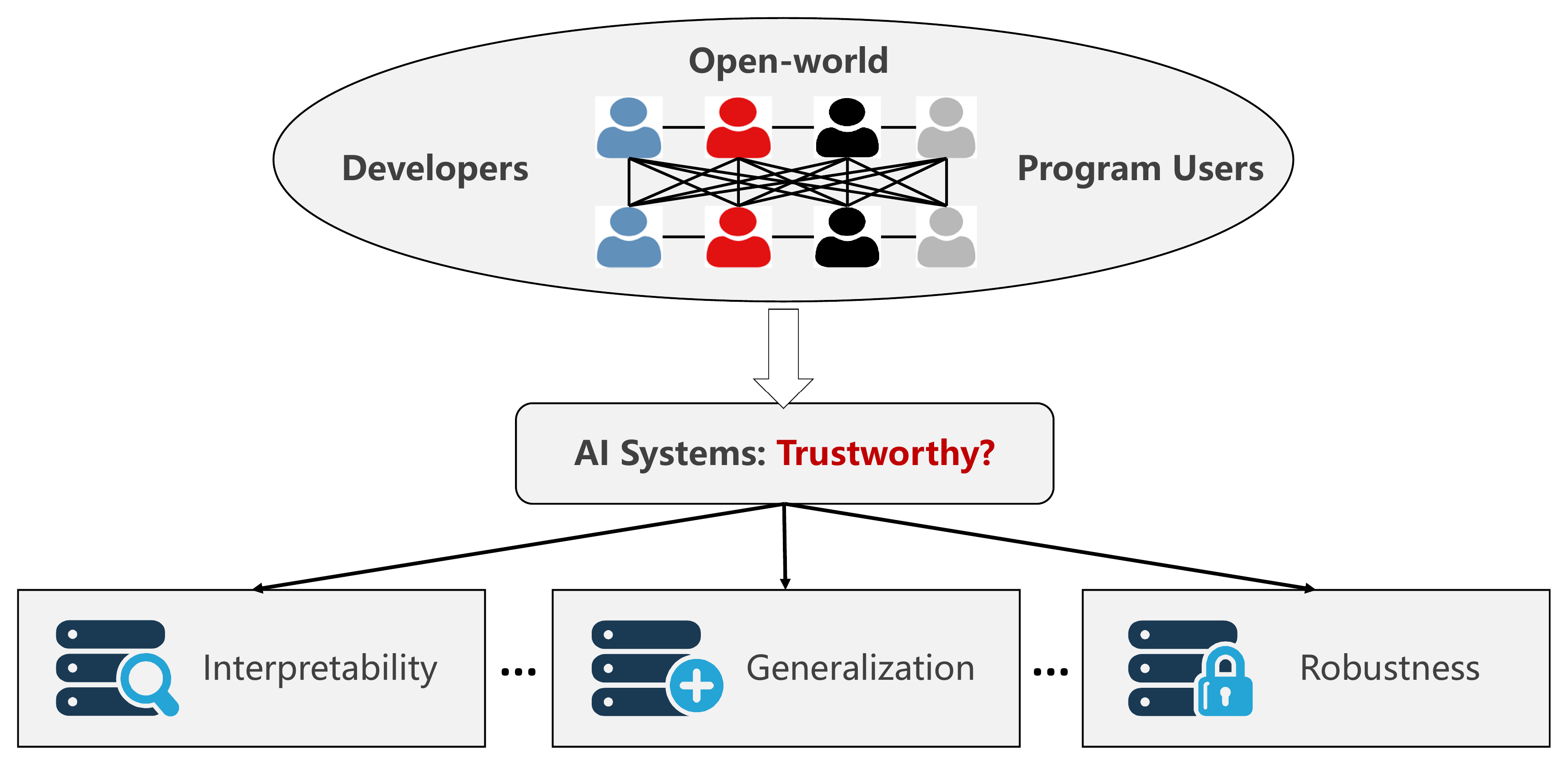}\\
  \caption{Several dimensions of the trustworthy AI systems could to be enhanced.}
  \label{Framework0}
\end{figure}

Contemporary artificial intelligence (AI) continues to furnish benefits to real-society from economic and environmental perspectives, among others \cite{Vinuesa2020Role, Goh2021Artificial}.
As AI gradually penetrates into high-risk fields such as healthcare, finance and medicine, which are closely related to human attributes, there is growing consensus awareness that people urgently expect these AI solutions to be trustworthy \cite{Kaur2022Trustworthy, Fei2022Towards}.
For instance, lenders expect the system to provide credible explanations for rejecting their applications; engineers wish to develop common system interfaces to adapt to wider environments; businesspeople desire that the system can still operate effectively under various complex conditions, among other expectations.
The rapid proliferation of these AI solutions has resulted in a crisis of trust, as the destruction of the trust of co-interest holders may have serious social consequences.
Such crises can include: 1) Prediction is difficult to understand; 2) Poor generalization ability; 3) Sensitivity to abnormal environments.
There can be summarized as the trustworthiness crisis of AI systems in terms of interpretability \cite{Zhang2022Protgnn, Xie2022Optimization, Li22Optimization}, generalization \cite{Zhang2020Generalized, Li2022OOD-GNN, Zhou2023Domain} and robustness \cite{Sun20Adversarial, Boult21Towards, Li2023Trustworthy}.
In stark contrast, professional AI developers have conventionally emphasized significant model performance (e.g., accuracy) as the ultimate criterion for their workflow.
From the perspective of ordinary and non-specialized AI beneficiaries, this metric is far from convincingly reflecting the trustworthiness of these AI systems.
To this end, it is imminent to construct trustworthy AI systems that go beyond performance, considering but not limited to the properties shown in Figure \ref{Framework0} to alleviate the trustworthiness crisis.

Against this backdrop, the exploration of trustworthy learning towards open-world has emerged as a fashionable tendency among researchers as one of the significant implementation methods for AI systems.
Henceforth, enormous amounts of developers have been working on concluding a general solution for constructing trustworthy frameworks \cite{Liu2022Trustworthy, Li2023Trustworthy, Wang2022GCL} (kindly refer to Section 1 of the \textbf{Appendix} for more related work on trustworthiness).
Concomitantly, the development of trustworthy learning also encounters the same three challenges as trustworthy AI that need to be bridged: 1) The dilemma of inadequate interpretability arising from network opaqueness properties; 2) The problem of deficient generalization attributed to restricted model cognitive capabilities; 3) The limitation of insufficient robustness caused by various unknown data instances.
Consequently, it is hardpressed to construct a trustworthy open-world learning for boosting the development of AI.

\begin{figure*}[!htbp]
  \centering
  \includegraphics[width=\textwidth]{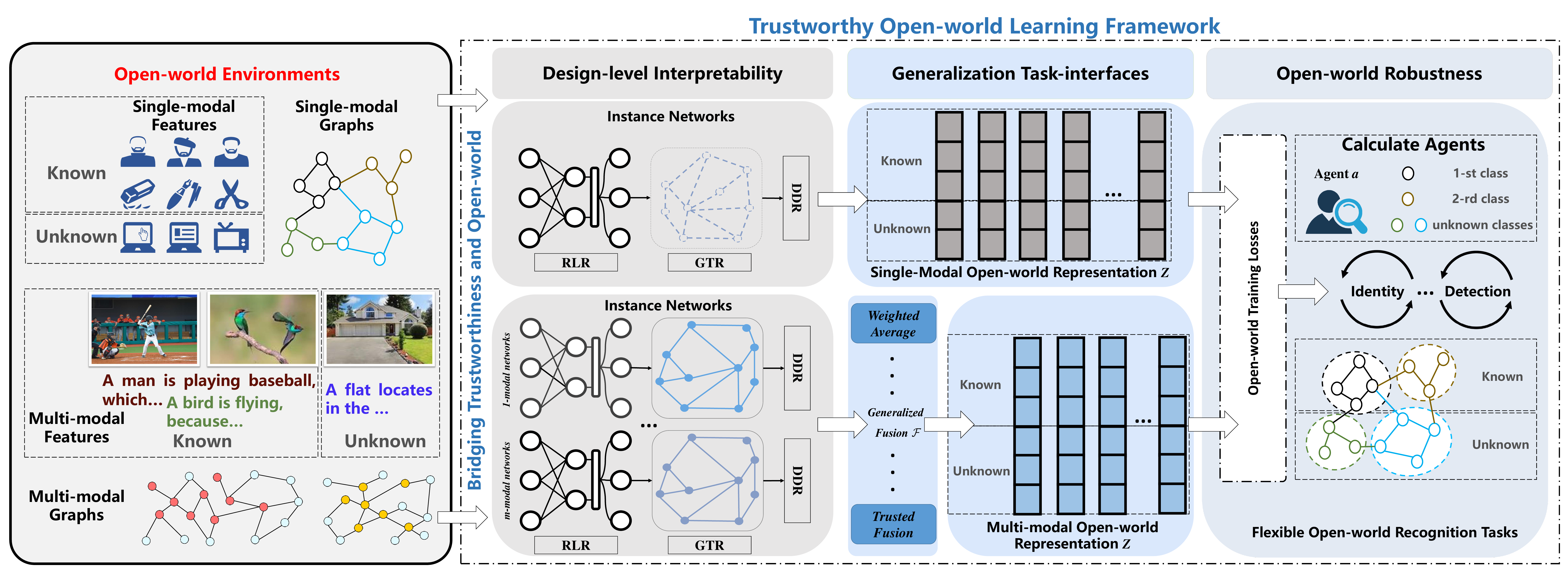}\\
  \caption{An overview of the proposed trustworthy open-world learning frameworks.}
  \label{Framework}
\end{figure*}

To alleviate the limitations and drawbacks mentioned above, this paper bridges trustworthiness and open-world learning for formally defining a family of design neural programs of trustworthy learning.
In order to achieve the purpose of this paper, we mainly fulfill it through the following steps:
1) Derived from a unified class of optimization problem-derived technologies, objective functions with specific physical meanings facilitate to construct trustworthy networks with the design-level interpretability;
2) For the pursuit of increasing the generalization of trustworthy learning, the environmental well-being task-interfaces boost the perception capabilities of models with flexible representation learning regularizers (RLR), demand-driven regularizers (DDR) and graph-topological regularizers (GTR);
3) Meticulously-designed open-world recognition losses and agent selection mechanisms assist the models in handling unknown or out-of-distribution inputs effortlessly, thus promoting the robustness of trustworthy learning.
Naturally, we maintain the above-mentioned trustworthiness properties to a more generalized multi-modal scenario.
This is accomplished by proposing a comprehensive generalizeable trustworthy protocol capacity of enhancing trustworthy properties in an operational setting.
The overall framework is demonstrated in Figure \ref{Framework}.
Accordingly, we expect to furnish the readers with enlightments into constructing such a congener trustworthy framework.
In a nutshell, the main contributions of this paper can be listed as follows:
\begin{itemize}
\item Bridge trustworthiness and open-world learning by exploring a family of neural approaches for enhancing interpretability, generalization, and robustness.
To the best of our knowledge, this is the first work on bridging trustworthiness and open-world learning, and enhancing trustworthy properties.
\item Establish interlinkages of trustworthy properties with the design-level explainability, environmental well-being task-interfaces and open-world recognition programs.
\item Provide guideline in an operational setting for these full-interested developers and expedite correlative researches on these essential trustworthy problems and solutions.
\end{itemize}

\section{The Proposed Framework}\label{sec:the}

In this section, we describe the procedure for designing trustworthy learning frameworks with enhancing trustworthiness properties.
From a single-modal perspective, we would explain how to increase the interpretability by optimization-inspired technologies at the design-level, rather than post-hoc explainability.
Afterwards, well-being task-interfaces with flexible learning regularizers boost the models empower their perception capabilities to hold their model-generalization.
Following closely, open-world recognition programs provide robustness for frameworks with handling out-of-distribution inputs effortlessly.
Grounded on this, by revisiting the above methods, more all-encompassing multi-modal scenario frameworks are maintained from a universal perspective for ensuring trustworthiness.
Hereto, we accomplish this by proposing a generalizeable trustworthiness learning capacity of constructing various trustworthiness properties in an operational setting.

\textbf{Notations.} Suppose that $\mathbf{X} = [\mathbf{X}^{ij}]_{N \times D}$ contains mixed known and unknown data, wherein $N$ is the total number of samples, and $D$ is the dimensionality of the distinct features.
$\mathbf{S}=[\mathbf{S}^{ij}]_{N \times N}$ denotes the similarity among samples.
$\mathbf{L}=[\mathbf{L}^{ij}]_{N \times N}$ marks a Laplacian matrix, computed as $\mathbf{L}=\mathbf{E}-\mathbf{S}$, where $\mathbf{E}$ is the degree matrix of $\mathbf{S}$.
$\mathbf{Z}=[\mathbf{Z}^{ij}]_{N \times K}$ indicates an open-world latent representation, where $K$ is the number of classes of known samples.
It should be noted that all bold and lowercase letters correspond to the generalized abstract form of the relevant matrix, and all data are normalized.

\subsection{Machine Optimization-based Design-level Interpretability}\label{sec:opt}

In pursuit of achieving the design-level interpretability, it is necessary to introduce some general terms with optimization that have a physical meaning, which prompts us to reassess fundamental physical concepts and their meanings.
Normally, upon receiving real-world data with various types of information, it is customary for it to possess high dimensionality.
Thus, it is rational to extract significant characteristics of the relationship among the data.
For example, in signal processing, we can leverage this ability to acquire knowledge of the signal and thus reconstruct the original signal to the maximum extent possible.
%Furthermore, in machine learning, we can utilize this approach to gain an understanding of the relationships between the various features of the data, as well as their corresponding underlying representative features.
In physics, particularly in quantum mechanics and condensed matter physics, this perspective can be interpreted as a measurement of the energy or cost associated with transforming the matrix into the change of basis or a quantum mechanical evolution \cite{Schutt19Unifying}.
Therefore, it is critical to consider it as a cornerstone in the explainable network design, and the above optimization scenario can be summarized as
\begin{equation}\label{Problem1}
M_f(\mathbf{x}):=\min_{\mathbf{z}}f(\mathbf{z}),
\end{equation}
where $M_f(\mathbf{x}): \mathbb{R}^{n} \rightarrow \mathbb{R}$ means a matrix approximation term that has a physical meaning and can encompass various interpretable application scenarios, among others, such as $\|\mathbf{X}-\mathbf{Z}\mathbf{D}\|_{F}^{2}$ (latent representation learning \cite{GregorL10}), $\|\mathbf{X}-\mathbf{X}\mathbf{Z}\|_{F}^{2}$ (subspace learning \cite{Wang20Smoothness}, $\mathbf{Z}=[\mathbf{Z}^{ij}]_{N \times N}$) and so forth.
However, as previously noted, due to the numerous dimensions involved in machine decision-making, this can result in an unsatisfactory generalization ability of the model when inspired by Problem \eqref{Problem1}.
Consequently, it becomes imperative to introduce a learning term that serves with the physical significance of balancing the model complexity with its ability to generalize to new data, especially for unknown data that requires decision-making.
Specifically, Problem \eqref{Problem1} can be rewritten as
\begin{equation}\label{Problem2}
M_g(\mathbf{x}):= M_f(\mathbf{x})+g(\mathbf{z}),
\end{equation}
where $g(\mathbf{z}): \mathbb{R}^{n} \rightarrow \mathbb{R}$ can encourage the model to simplify the open-world representation of its input data by focusing only on the most relevant features and ignoring noise or irrelevant information, such as sparse-norm, nuclear-norm, and their derivatives.
Typically, a possible tighter surrogate of the objective function in \eqref{Problem2} is to guarantee $M_f(\mathbf{x})$ and relaxes $g(\mathbf{z})$ only, providing a possible solution.
Therefore, the subdifferential, proximal operator and Moreau envelope of a proper convex function $M_g(\mathbf{x})$ \cite{BeckTeboulle09AFast} are defined as
\begin{equation}\label{Problem3}
\left\{\begin{array}{l}
\partial f(\mathbf{x}):=\{\mathbf{a} \in \mathcal{E}: f(\mathbf{y}) \geq f(\mathbf{x})+\langle\mathbf{y}-\mathbf{x}, \mathbf{a}\rangle, \forall \mathbf{y} \in \mathcal{E}\} \\
\operatorname{prox}_{\mu \cdot g}(\mathbf{x}):=\left\{\mathbf{z} \in \mathcal{E}: \mathbf{z}=\underset{\mathbf{z}}{\operatorname{argmin}} \frac{1}{2 \mu}\|\mathbf{z}-\mathbf{x}\|^2+g(\mathbf{z})\right\} \\
M_g^\mu(\mathbf{x}):=\underset{\mathbf{z}}\min \frac{1}{2 \mu}\|\mathbf{z}-\mathbf{x}\|^2+g(\mathbf{z}),
\end{array}\right.
\end{equation}
where $\operatorname{prox}(\cdot)$ is a proximal operation $\mu > L(f)$, $L(f)$ is the Lipschitz constant of $f(\cdot)$, and $\mathcal{E}$ is the Euclidean space.
Actually, the sparse-norm, nuclear-norm, and their derivatives all necessitate proximal operations, although the procedures executed to satisfy their constraints are distinct.
Therefore, with the help of demand-driven regularizer \eqref{Problem3}, we can integrate a generalized solution to Problem \eqref{Problem2} with physical implications, equivalently
\begin{equation}
\begin{array}{ll}\label{Z_ISAT}
\mathbf{z}^{\left( t+1 \right)} \leftarrow \operatorname{prox}_{g}\left( \mathbf{z}^{\left( t \right)}-\frac{1}{L}\nabla f(\mathbf{z}^{\left( t\right)} ) \right),
\end{array}
\end{equation}
where $\nabla f(\cdot)$ is the gradient of Problem \eqref{Problem1}, and $t$ is the $t$-th iteration.
The demand-driven regularizer \eqref{Z_ISAT} extends some prior techniques \cite{BeckTeboulle09AFast, Liu10Robust} and its physical significance lies in its capacity to identify the solution that simultaneously satisfies Problem \eqref{Problem2}.
Further, we can formalize the interpretable machine optimization equation
\begin{equation}\label{SolvingZ}
\underbrace{{\mathbf{z}}^{(t+1)}\leftarrow\operatorname{prox}_{g}\left( \mathbf{z}^{\left( t \right)}, \mathbf{x} \right).}_{\textbf{Machine iterative equation with interpretability}}
\end{equation}
Machine optimization-based Problem \eqref{Problem2} and regularizer \eqref{Z_ISAT} provide a solid foundation \eqref{SolvingZ} for moving towards network design-level interpretability.
Furthermore, we will advance towards constructing trustworthy frameworks that are explainable grounded on these machine optimization techniques.

%\begin{figure}[t]
%  \centering
%  \includegraphics[width=0.48\textwidth]{./Figures/Framework_acmmm_2.pdf}\\
%  \caption{Step design-level interpretability networks.}
%  \label{Framework2}
%\end{figure}

\subsection{Step Forward: Generalization Environmental Well-being Task-interfaces}\label{sec:gen}

With the aid of these interpretability tools that are imbued with physical significance, we will now proceed towards deep networks.
Currently, the groundwork for interpretability has been established in \eqref{SolvingZ}, but it is limited to the optimized architecture level.
In another word, there exist notable obstacles that impede our progress towards deep networks, particularly the fact that current schemes lack the same implicit optimization parameters in \eqref{SolvingZ} as deep networks.
In addition, in machine optimization learning, a long-standing issue is manually tuning hyper-parameters such as the coefficient in front of $g(\cdot)$, which considerably hinders the working efficiency.
To overcome these limitations and advance towards deep networks, we utilize measures such as adding learnable variables by deduction and substitution in the demand-driven regularizer \eqref{Z_ISAT}.
Moreover, we also enable these frameworks to automatically search for an appropriate balance value by parameter self-learning, expressed as
\begin{equation}\label{SolvingZ_ista}
\underbrace{{\mathbf{z}}^{(t+1)}\leftarrow\mathcal{P}_{\theta}\left( \mathbf{z}^{\left( t \right)}, \mathbf{x}; \mathbf{\Theta}_{\mathbf{z}} \right),}_{\textbf{Step forward: Demand-driven regularizer network layers}}
\end{equation}
where $\mathcal{P}_{\theta}$ in \eqref{SolvingZ_ista} is reparameterized with $\operatorname{{prox}}_{g}(\cdot)$, which facilitates the efficient automatic search of task-specific hyper-parameter $\theta$, and $\mathbf{\Theta}_{\mathbf{z}}$ is a learnable parameter.
Utilizing the network layer \eqref{SolvingZ_ista}, we transition from machine optimization to constructing a generalized demand-driven regularizer-centered task-interface while preserving interpretability (as outlined in Subsection \ref{sec:opt}).
Although \eqref{SolvingZ_ista} is applicable to structured data, it is not as workable for unstructured ones, such as social networks data.
To address this, we introduce a graph-topological regularizer term that provides a more versatile task-interface as
\begin{equation}\label{SolvingZ_ista_Graph}
\underbrace{{\mathbf{z}}^{(t+1)}\leftarrow\mathcal{P}_{\theta}\left( \mathbf{z}^{\left( t \right)}, \mathbf{x}, \boldsymbol{h}; \mathbf{\Theta}_{\mathbf{z}} \right),}_{\textbf{More generalized: Graph-topological regularizer network layers}}
\end{equation}
where the graph-topological regularizer $\boldsymbol{h}(\cdot)$ encourages the model to learn a smoothness and continuous open-world representation $\mathbf{z}$.
Herein, Equation \eqref{SolvingZ_ista_Graph} is associated with problem $M_h(\mathbf{x}):= M_f(\mathbf{x})+g(\mathbf{z})+h(\mathbf{z})$.
Its physically considers the graph-topological structure of the data, and also has good interpretability.
Thus far, we have utilized the design-level interpretability provided by machine optimization to shift towards deep networks and further established two generalization environmental well-being task-interfaces \eqref{SolvingZ_ista}-\eqref{SolvingZ_ista_Graph} to enhance these trustworthy properties.

\subsection{Towards Open-world Robustness}\label{sec:tow}

\begin{figure}[t]
  \centering
  \includegraphics[width=0.48\textwidth]{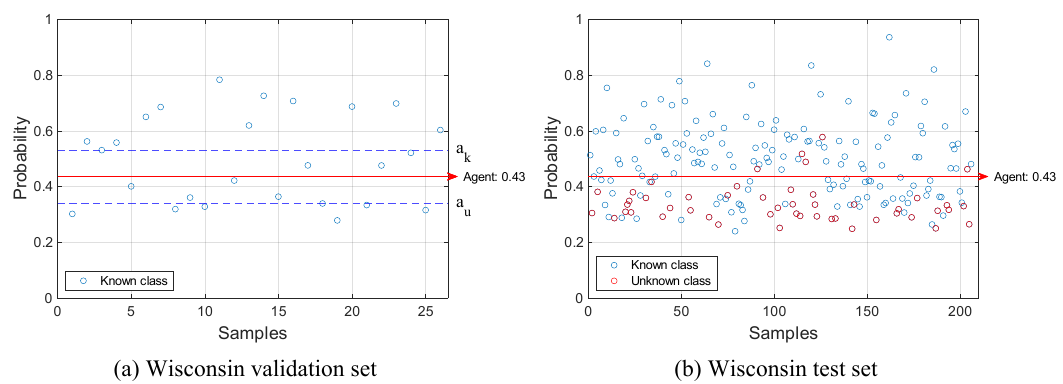}\\
  \caption{A visualization of determining the agent using a validation set and employing it to test.}
  \label{Framework3}
\end{figure}

In the practical realm, models are required not only to handle familiar data but also to operate in an open-world setting, where known and unknown data are typically intertwined for decision-making.
%The open-world work in operational phases, each of which is composed of an uncontrolled real-world environment.
Indeed, we derive the representation using Frameworks \eqref{SolvingZ_ista}-\eqref{SolvingZ_ista_Graph} in an open-world environment that encompasses two components: representation for known and unknown data.
To enhance the robustness of the previous trustworthy frameworks and adapt them for open-world scenarios, we have devised the ensuing losses
\begin{equation}
\begin{array}{ll}\label{lossCla1}
\mathcal{L}_{k} = -\frac{1}{N_{k}}\sum\limits_{i=1}^{N_{k}} \sum\limits_{j=1}^K \mathbf{Z}^{ij}_{k} \log \hat{\mathbf{Z}}^{ij},
\end{array}
\end{equation}
\begin{equation}
\begin{array}{ll}\label{lossCla2}
\mathcal{L}_{u} = \frac{1}{N_{u}}\sum\limits_{i=1}^{N_{u}} \sum\limits_{j=1}^K \mathbf{Z}^{ij}_{u} \log \hat{\mathbf{Z}}^{ij},
\end{array}
\end{equation}
\begin{equation}
\begin{array}{ll}\label{lossCla}
&\mathcal{L}_{total} = \lambda_{1}\mathcal{L}_{k}+\lambda_{2}\mathcal{L}_{u},
\end{array}
\end{equation}
where $\mathbf{Z}^{ij}_{k}$ and $\mathbf{Z}^{ij}_{u}$ denote one-hot coding of ground-truths and pseudo labels of unlabeled samples through $argmax(\mathbf{Z}_{u})$, respectively.
$\hat{\mathbf{Z}}^{ij}$ indicates the set probability of known and unknown data that the $i$-th representative sample belongs to class $j$, and $\lambda_{1}$ and $\lambda_{2}$ are two trade-off balance parameters.
It should be noted that we do not minimize loss $\mathcal{L}_{u}$ (removing minus sign), as our objective is to attain recognition by maximizing the uncertainty of the unknown classes.
Furthermore, we rank the normalized open-world representations and discard samples whose ranking values fall within the bottom and top 10$\%$, respectively.
This is because large values are favorable for recognition, while small values indicate a balanced output across each known class, which is more conducive to enhancing the model robustness.
%Through training with loss function \eqref{lossCla}, we aim to increase the discriminatory power of the recognized classes by labeling the data, while simultaneously maximizing the uncertainty loss to achieve a more balanced output for each sample, which assists in detecting unknown classes.
Additionally, to improve the model capability for open-world recognition, we incorporate an agent $a$ into the perception to aid in identifying whether a sample belongs to a known or an unknown class as
\begin{equation}\label{Reject}
\hat{y}_{i}=\left\{\begin{array}{cc}
\operatorname{Unknown}, & \text { if } \max _{k \in K} p\left(k \mid x_i\right) \leq a, \\
\arg \max _{k \in K} p\left(k \mid x_i\right), & \text { otherwise, }
\end{array}\right.
\end{equation}
where $p\left(k \mid x_i\right)$ is obtained from the softmax output of representations, and $\hat{y}_{i}$ is the prediction label.
If the value of the present sample is below the agent $a$, we reject it as an instance of an unknown class; otherwise, we designate the predicted class as the one with the highest probability \cite{Wu20OpenWGL}.
Figures \ref{Framework3}-\ref{Framework4} vividly depict the above process.
Please refer to Subsections 2.3 and 2.4 of the \textbf{Appendix} for more open-world losses and agent selection details.

\subsection{Revisiting Overall Trustworthy Frameworks and Derived Examples}\label{sec:rev}

%Overall, we have previously outlined the measures taken to enhance interpretability, generalization, and robustness in the proposed framework.
%Specifically, the framework incorporates physically meaningful machine optimizations to achieve design-level interpretability, which in turn facilitates a more generalized task network interface.
%Furthermore, we utilize carefully crafted loss and agent settings to improve the robustness of the model in an open-world setting.
When revisiting the proposed trustworthy frameworks, we observe that the output of the design-level interpretability networks is exactly the minimizer of an objective function during model training.
This is particularly interesting because it aligns with the concept of bi-level optimization \cite{Colson07Overview} protocols, which can be formalized as
\begin{equation}\label{BiProblem}
\begin{aligned}
&\min_{\mathbf{\theta}} \mathcal{L}_{total}(\mathbf{z}^{*}(\mathbf{x}, \mathbf{\theta}), \mathbf{y}), \\ \textbf{s.t. } \mathbf{z}^{*}(\mathbf{x}, \mathbf{\theta}) & \in \arg\min_{\mathbf{z}} M_h(\mathbf{x}): M_f(\mathbf{x})+g(\mathbf{z})+h(\mathbf{z}),
\end{aligned}
\end{equation}
where the lower-level regularizer-centered networks are designed to obtain a meaningful representation $\mathbf{z}^{*}$, and $\mathbf{y}$ is the ground-truths, while the upper-level open-world training losses are incorporated to enhance the awareness of task representations by optimizing the model parameter $\mathbf{\theta}$.

\begin{figure}[t]
  \centering
  \includegraphics[width=0.48\textwidth]{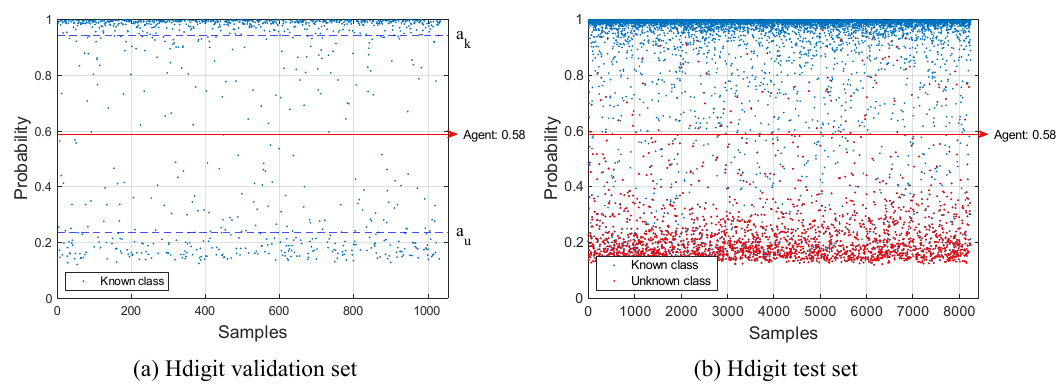}\\
  \caption{A visualization of determining the agent using a validation set and employing it to test.}
  \label{Framework4}
\end{figure}

Under the supervision of the above protocols, we reorganize the exploration and design process of the entire proposed trustworthy framework.
Initially, we confer interpretability to the trustworthy frameworks at the design-level grounded on open-world data physics principles and machine optimization.
Subsequently, with the assistance of machine-optimized regulations, we progress towards deep networks and establish generalization for dependable frameworks at the task-level.
Concurrently, the model capability to discern unknown entities is augmented via the utilization of open-world losses and agent rules with the data-level robustness.
The above frameworks can be summarized as Protocol (Algorithm) 1 of the \textbf{Appendix}, and some specific examples are given below.

\subsubsection{Demand-driven Regularizer-centered Network Layers}

When reviewing the demand-driven regularizer-centered Framework \eqref{SolvingZ_ista}, several examples can be derived, as outlined in Table 1 of the \textbf{Appendix}, which can be generalized under Framework \eqref{SolvingZ_ista} as follows
\begin{equation}\label{SolvingZD}
\underbrace{{\mathbf{Z}}^{(t+1)}\leftarrow\mathcal{P}_{\theta}\Big(\mathbf{Z}^{(t)}\mathbf{F} + H(\mathbf{X}, \mathbf{U})\Big),}_{\textbf{{Instantiated network layers in generalized Framework \eqref{SolvingZ_ista}}}}
\end{equation}
where $\mathcal{P}_{\theta}$ means a generalized demand-driven regularizer, $H(\cdot, \cdot)$ denotes a residual term, and $\mathbf{F}$, $\mathbf{U}$ are learnable parameters.
Instantiated generalized equation \eqref{SolvingZD} results in residual-guided demand-driven networks, which consist of a representation learning term, a data residual term, and a demand-driven activation function.

\subsubsection{Graph-topological Regularizer-centered Network Layers}

While the above framework meets graph-topological regularizers, it can collide with Framework \eqref{SolvingZ_ista_Graph}.
Several examples can also be derived as outlined in Table 1 of the \textbf{Appendix}, which can be generalized under Framework \eqref{SolvingZ_ista_Graph} as
\begin{equation}\label{SolvingZG}
\underbrace{{\mathbf{Z}}^{(t+1)}\leftarrow\mathcal{P}_{\theta}\Big(\mathbf{Z}^{(t)}\mathbf{F} - \alpha G(\mathbf{L})\mathbf{Z}^{(t)}\mathbf{W} + H(\mathbf{X}, \mathbf{U})\Big),}_{\textbf{{Instantiated network layers in generalized Framework \eqref{SolvingZ_ista_Graph}}}}
\end{equation}
where $G(\mathbf{L})\mathbf{Z}^{(t)}\mathbf{W}$ gives a generalized graph item, $\alpha$ means a contraction factor, and $\mathbf{W}$ is a learnable parameter.
Instantiated equation \eqref{SolvingZG} results in graph-topological demand-driven layers, which consist of a representation learning term, a graph regularizer term, a data residual term, and a demand-driven function.

\subsection{Beyond Single-modal: Trustworthy Multi-modal Framework From a Universal Perspective}\label{sec:bey}

%The domain of artificial intelligence has accomplished remarkable advancements in recent years.
%Notably, one area that has witnessed considerable progress is the advancement of multi-modal AI systems.
%Prior to the evolution of multi-modal AI, the majority of AI systems are single-modal, signifying that they could only process and analyze data from a single source, such as text or images.
%Despite being valuable in numerous applications, these systems have limitations in their ability to comprehend intricate and subtle situations that require multiple sources of information.
%Compared with the single-model AI, a crucial advantage of multi-modal ones is that it can provide a more comprehensive understanding of a situation or problem, thus improving the performance of existing AI systems.
%Motivated by these developments, we build a multi-modal trustworthy framework that beyond a aforementioned single-modal ones from a universal perspective.

\textbf{Notations.} Suppose there are $M$ modalities data, denoted as $\{\mathbf{X}_{m}\}_{m=1}^{M}$, which include mixed known and unknown data in the $m$-th modality.
The similarity among samples in the $m$-th modality is expressed by $\mathbf{S}_{m}=[\mathbf{S}_{m}^{ij}]_{N \times N}$.
The Laplacian matrix, marked by $\mathbf{L}_{m}=[\mathbf{L}_{m}^{ij}]_{N \times N}$, is computed as $\mathbf{L}_{m}=\mathbf{E}_{m}-\mathbf{S}_{m}$, where $\mathbf{E}_{m}$ is the degree matrix of $\mathbf{S}_{m}$.
The $m$-th modality latent representation is indicated by $\mathbf{Z}_{m}=[\mathbf{Z}_{m}^{ij}]_{N \times K}$, and a multi-modal open-world co-latent representation is denoted by $\mathbf{Z}$.
%It should be noted that all bold and lowercase letters also correspond to the vector form of the matrix, and all data are normalized.

Likewise, it is imperative to model the problem in multi-modal scenarios like \eqref{Problem2}, and then use machine optimization rules to expand it.
In this regard, we can directly extend Framework \eqref{SolvingZ_ista} to a multi-modal demand-driven task-interface by learning the multi-modal open-world co-latent representation $\mathbf{z}$ as
\begin{equation}\label{Multi_SolvingZ_ista}
\underbrace{{\mathbf{z}}^{(t+1)}\leftarrow\mathcal{F}\left(\mathcal{P}_{\theta_{m}}( \mathbf{z}_{m}^{\left( t \right)}, \mathbf{x}_{m}; \mathbf{\Theta}_{\mathbf{z}_{m}}); \mathbf{\Theta}_{\mathcal{F}}\right),}_{\textbf{Multi-modal demand-driven regularizer network layers}}
\end{equation}
where $\mathcal{F}$ denotes a generalized fusion, which can be weighted average $\mathcal{V}$, auto-weight fusion $\mathcal{W}$, attention mechanism $\mathcal{A}$ \cite{Vaswani17Attention} and trusted fusion $\mathcal{T}$ \cite{Han2023Trusted}, whereas $\mathbf{\Theta}_{\mathbf{z}_{m}}$ denotes the $m$-th learnable parameter.
Similarly, Framework \eqref{SolvingZ_ista_Graph} can also be expanded to a multi-modal graph-topological task-interface as
\begin{equation}\label{Multi_SolvingZ_ista_Graph}
\underbrace{{\mathbf{z}}^{(t+1)}\leftarrow\mathcal{F}\left(\mathcal{P}_{\theta_{m}}( \mathbf{z}_{m}^{\left( t \right)}, \mathbf{x}_{m}, \boldsymbol{h}_{m}; \mathbf{\Theta}_{\mathbf{z}_{m}} ); \mathbf{\Theta}_{\mathcal{F}}\right).}_{\textbf{Multi-modal graph-topological regularizer network layers}}
\end{equation}
Herein, we have transitioned from Frameworks \eqref{SolvingZ_ista}-\eqref{SolvingZ_ista_Graph} to \eqref{Multi_SolvingZ_ista}-\eqref{Multi_SolvingZ_ista_Graph}, resulting in a unversial multi-modal trustworthy framework with a single-modal foundation, significantly enhancing the user experience.
Furthermore, the above frameworks enhance design-level interpretability in multi-modal scenarios, while also increases generalization and robustness.
Here, we can also formalize the above opinion into the following multi-modal bi-level optimization framework \cite{Colson07Overview} as
\begin{equation}\label{MultiBiProblem}
\begin{aligned}
&\min_{\mathbf{\theta}_{m}, \mathbf{\theta}_{\mathcal{F}}} \mathcal{L}_{total}\Big(\mathcal{F}\left(\mathbf{z}_{m}^{*}(\mathbf{x}_{m}, \mathbf{\theta}_{m}), \mathbf{\theta}_{\mathcal{F}}\right), \mathbf{y}\Big), \\ \textbf{s.t. } \mathbf{z}_{m}^{*}(\mathbf{x}_{m}, &\mathbf{\theta}_{m})  \in\arg\min_{\mathbf{z}_{m}} \sum_{m=1}^{M} M_h(\mathbf{x}_{m}): M_f(\mathbf{x}_{m})+g(\mathbf{z}_{m})+h(\mathbf{z}_{m}),
\end{aligned}
\end{equation}
where the upper-level open-world training losses and the lower-level multi-modal regularizer-centered networks together constitute the above frameworks, as shown in Protocol (Algorithm) 2 of the \textbf{Appendix}, and several examples are provided below.

\subsubsection{Multi-modal Demand-driven Regularizer-centered Network Layers}
In multi-modal scenarios, we can generalize some examples in Table 3 of the \textbf{Appendix} under Framework \eqref{Multi_SolvingZ_ista} as follows
\begin{equation}\label{SolvingZDM}
\underbrace{{\mathbf{Z}}_{m}^{(t+1)}\leftarrow\mathcal{P}_{\theta_{m}}\Big(\mathbf{Z}_{m}^{(t)}\mathbf{F} + H(\mathbf{X}_{m}, \mathbf{U}_{m})\Big),}_{\textbf{{Instantiated multi-modal network layers in generalized Framework \eqref{Multi_SolvingZ_ista}}}}
\end{equation}
where network layer \eqref{SolvingZDM} is the multi-modal version of \eqref{SolvingZD}.
In this case, the obtained representation group $\mathbf{Z}_{m}^{(t)}$ needs to be generalized fusion, which is denoted as follow
\begin{equation}\label{SolvingZFusion}
\underbrace{{\mathbf{Z}}^{(t+1)} \leftarrow \mathcal{F}({\mathbf{Z}}_{1}^{(t+1)}, \cdots, {\mathbf{Z}}_{m}^{(t+1)}; \mathbf{\Theta}_{\mathcal{F}}), }_{\textbf{{Multi-modal generalized fusion in Frameworks \eqref{Multi_SolvingZ_ista}-\eqref{Multi_SolvingZ_ista_Graph}}}}
\end{equation}
where $\mathbf{\Theta}_{\mathcal{F}}$ is a learnable parameter in generalized fusion.

\subsubsection{Multi-modal Graph-topological Regularizer-centered Network Layers}
When we consider multi-modal graph-topological regularizer-centered situations, examples in Table 3 of the \textbf{Appendix} can also be included under Framework \eqref{Multi_SolvingZ_ista_Graph}, as shown below
\begin{equation}\label{SolvingZGM}
\underbrace{{\mathbf{Z}}_{m}^{(t+1)}\leftarrow\mathcal{P}_{\theta_{m}}\Big(\mathbf{Z}_{m}^{(t)}\mathbf{F} - \alpha G(\mathbf{L}_{m})\mathbf{Z}_{m}^{(t)}\mathbf{W} + H(\mathbf{X}_{m}, \mathbf{U}_{m})\Big),}_{\textbf{{Instantiated multi-modal network layers in generalized Framework \eqref{Multi_SolvingZ_ista_Graph}}}}
\end{equation}
where network layer \eqref{SolvingZGM} is the multi-modal version of \eqref{SolvingZG}, and a generalized fusion is also required.
Overall, the transition from a single-modal trustworthy framework to multi-modal ones is a significant stride forward in the advancement of more intricate and effective AI systems.
%\textbf{Please refer to the Appendix for discussions with existing work.}

\subsection{Discussions and Insights}\label{sec:dis}

\subsubsection{Relationships with Previous Works}\label{sec:pre}

In this subsection, we explore the possibility that some previous excellent works can be incorporated into the proposed frameworks, while maintain these trustworthy properties.

% \cite{Gu2020Implicit, Liu21EIGNN, Liu2022MGNNI}
\textbf{Connection with Implicit Networks:} As proposed in excellent works such as \cite{Liu21EIGNN, Liu2022MGNNI}, various implicit networks define a fixed-point equation as an implicit layer for aggregation, thereby generating the equilibrium representation.
In this context, the proposed frameworks can harness the following transformations to build connections to implicit networks as
\begin{equation}\label{SolvingINN}
\begin{aligned}
&\min_{\mathbf{\theta}}\mathcal{L}(\mathbf{z}^{*}(\mathbf{x}, \mathbf{\theta}), \mathbf{y}), \textbf{s.t. }\\&\mathbf{z}^{*}(\mathbf{x}, \mathbf{\theta}) \in \arg\min_{\mathbf{z}} M_h(\mathbf{x}),
\end{aligned} \quad \Leftrightarrow \quad
\begin{aligned}
&\min_{\mathbf{\theta}}\mathcal{L}(\mathbf{z}^{*}(\mathbf{x}, \mathbf{\theta}), \mathbf{y}), \textbf{s.t. }\\&\mathbf{z}^{*}(\mathbf{x}, \mathbf{\theta}) \in \operatorname{Fix} (\mathcal{B}(\mathbf{x}, \mathbf{\theta})),
\end{aligned}
\end{equation}
where $\operatorname{Fix} (\mathcal{B}(\mathbf{x}, \mathbf{\theta}))$ is a fixed-point equation, which can be derived from the proposed framework.
For example, the implicit framework in \cite{Liu2022MGNNI} is a special case of generalized network layer (14) ($\mathbf{F}$ is equal to zero and the minus sign before $G(\cdot, \cdot)$ is placed in learnable parameters).
Consequently, these methods can be seamlessly integrated into the proposed frameworks ((6)-(7) to (15)-(16)) to create various types of trustworthy learning methods that maintain trustworthiness and enhance their diverse theorems and properties.

\textbf{Connection with Graph Convolutional Networks and Variants:} Leveraging the insights from some outstanding interpretable graph convolutional networks \cite{Zhu2021interpreting, Wang23Beyond}, the proposed trustworthy methods can broadly encompass some existing GCN approaches, as depicted in Table 2 of the \textbf{Appendix}.
Specifically, some of its variants (such as the connection with hypergraph neural network in Table 3 of the \textbf{Appendix}) can also be crafted and derived into the proposed frameworks.
The proposed trustworthy frameworks not only enhance interpretability as these works, but also further increase generalizability (generalized task-interfaces) and robustness (perception of unknown things) in open-world scenarios.

\textbf{Connection with Other Prior Networks:} The above-mentioned network layers can be further extended to multi-modal/view scenarios.
In addition, different from \cite{Wan2022Continual, Wang22Learning}, using traditional multivariate optimization methods such as alternating direction method of multipliers, the proposed frameworks can also be extended to multi-variate trustworthy network layers, where each subproblem can constitute a subnetwork.
Then, we can construct the corresponding trustworthy frameworks according to Protocols 1 or 2.
\textbf{It should be noted that most of the networks in Tables 1-3 of the \textbf{Appendix} are firstly proposed and derived from our frameworks, which also reflects the versatility and universality.}

Indeed, the aforementioned outstanding works (such as \cite{Zhu2021interpreting, Wang23Beyond}) may also explore a broader framework, but they all investigate a single property within a trustworthy framework, such as interpretability.
In contrast, this paper aims to enhance trustworthy properties and integrate them into a more comprehensive framework to bridge trustworthiness and open-world learning.

\subsubsection{Insight Remarks}\label{sec:rem}
Given the extensive research fields involved in each trustworthy property, we aim to provide readers with some insights in this regard through the superficial exploration of the frameworks presented in this paper.
Therefore, we provide the following observations on the proposed trustworthy frameworks.

\textbf{Remark 1: Trustworthy Reclaim.}
\begin{itemize}
\item Interpretability: The proposed trustworthiness protocol employs machine optimization rules to guide the development of deep networks that prioritize design-level interpretability, resulting in models that have physical meanings and are more easily understood by users.
\item Generalization: By incorporating representation learning, demand-driven and graph-topological regularizers into the design-level interpretability of the proposed trustworthy protocols, it can offer more adaptability and better support for generalized well-being interfaces and downstream tasks.
\item Robustness: By incorporating the proposed trustworthy framework into the open-world environments, it enables the model to identify unknown entities and improve the robustness.
\item Bridging trustworthiness and open-world learning: Integrating the aforementioned properties into Protocols 1 and 2 and following the subsequent steps to acquire a framework can enhance trustworthiness and improve the universality and user experience of these approaches in open-world.
    The proposed methods can occupy a niche in the trend of unifying both single-modal and multi-modal learning.
\end{itemize}

\textbf{Remark 2: Contribution Reclaim.}

Note that there are some of the methods that we utilize are existing works, such as \cite{GregorL10, Wu20OpenWGL, Vaswani17Attention, Han2023Trusted}, the combination of the proposed frameworks with these methods leads to a more generalized framework to alleviate the challenges of trustworthy learning.

\subsection{Theoretical Analysis}\label{sec:the}

\textbf{Convergence:} As a vital property of trustworthy learning, to prove their convergence, we insert the following \textbf{Theorem 1}.

%\textbf{Definition 1. (Banach fixed Point Theorem)} \textit{Let $(\mathbf{X}, d)$ be a non-empty complete metric space with a contraction mapping $T: \mathbf{X} \rightarrow \mathbf{X}$.
%Then, T admits a unique fixed-point $\mathbf{x}^{*}$ in $\mathbf{X}$ (i.e. $T(\mathbf{x}^{*})= \mathbf{x}^{*}$).
%Furthermore, $\mathbf{x}^{*}$ can be found as follows: start with an arbitrary element $\mathbf{x}_{0} \in \mathbf{X}$ and and define a sequence $\{\mathbf{x}_{n}\}_{n \in \mathbb{N}}$ by $ \mathbf{x}_{n} = T\{\mathbf{x}_{n-1}\}$ for $n \geq 1$. Then $\lim_{n \rightarrow \infty} \mathbf{x}_{n}=\mathbf{x}^{*}$.}

\textbf{Theorem 1.} \textit{Given the bounded damping factor $\alpha \in [0, 1)$, the proposed networks for propagation (such as Tables 1-3 of the \textbf{Appendix}) is a contraction mapping and the unique convergence solution $\mathbf{Z}^{*}$ can be obtained by the proposed frameworks.}

This can be proved by using the properties of matrix vectorization and the Kronecker product with the Banach fixed Point Theorem.
Please refer to Subsection 2.5 of the \textbf{Appendix} for proof.
About the proposed theorem, we have the following conclusions: 1) The models derived from the proposed framework can achieve optimal values on sufficient training rounds; 2) At least the models exported in this work can ensure convergence. 3) The convergence analysis in the experiment also verified this point.

%\textbf{Substitutability.} These parameterizations (such as $\mathbf{F}=\mathbf{I}-\frac{1}{L}\mathbf{D}_{m}^{T}\mathbf{D}_{m}$, $\mathbf{W}=\frac{1}{L}\mathbf{I}$ and $\mathbf{U}=\frac{1}{L}\mathbf{D}_{m}^{T}$ of the \textbf{Appendix}) are a linear compound operation that can be easily replaced.
%It is helpful to expand the solution space by replacing these operations with learnable parameter layers.

\textbf{Complexity:} The complexity of single-modal and multi-modal trustworthy learning frameworks requires $\mathcal{O}(N^{2}K)$ and $\mathcal{O}(MN^{2}K)$ for each epoch forward/backward propagation, respectively.

\section{Experiments}\label{sec:exp}

In order to put forward the proposal in this paper, several example networks were implemented using the proposed frameworks as the backbone to verify the rationality.
The following subsection provides a quantitative and qualitative analysis of the experimental results, including performance and trustworthiness.
Note that 10\% of the ground-truth labels are used for $\mathcal{L}_{k}$ training, while the pseudo labels generated by unlabeled samples are used for $\mathcal{L}_{u}$ training.
For a more specific introduction of experimental setups, including dataset details, compared methods and parameter settings, kindly refer to Section 3 of the \textbf{Appendix}.

\subsection{Experimental Results}\label{sec:res}
Since performance serves as a foundation for trustworthiness, we conducted experiments on both single-modal datasets and more complex multi-modal datasets for evaluation.
The experimental results and analysis are as follows.

\begin{table*}[t]
\centering
\caption{Accuracy of all compared single-modal semi-supervised classification methods under $10\%$ ratio of training labels and different unknown classes, where the best and running-up results are highlighted in bold (mean\%). OM means out-of-memory.}
\resizebox{\textwidth}{!}{
\begin{tabular}{c|c||cc|cc|cc|cccc|ccc||cc}
\toprule
\multicolumn{2}{c||}{Datasets $\backslash$ Methods} & MLP & AE & ASF-Net & WAST-Net & GCN & GAT & SGCN & FAGCN & AMGCN & HOGGCN & GNN-LF & GNN-HF & HGNN$^{+}$ & SL-Net & SGL-Net \\
\midrule
\multirow{2}*{Chameleon}       & Unknown = 1 & 23.48 & 22.67 & 24.43 & 30.98 & 30.35 & 29.76 & 31.16 & 34.69 & 35.69 & 22.64 & 34.20 & 34.31 & 33.18 & \textbf{37.14} & \textbf{38.98} \\
                             & Unknown = 2 & 25.63 & 27.98 & 28.43 & 33.57 & 36.14 & 45.56 & 48.87 & 50.69 & 47.87 & 42.09 & 42.51 & 40.09 & 38.11 & \textbf{53.76} & \textbf{54.02} \\
\midrule
\multirow{2}*{CoraFull}    & Unknown = 1  & 35.59 & 32.44 & 36.51 & 38.42 & 40.87 & OM & \textbf{46.97} & 41.01 & 43.10 & OM & \textbf{46.88} & 41.82 & 46.64 & 44.95 & 44.79 \\
                             & Unknown = 2 & 32.46 & 31.48 & 37.78 & 38.32 & 30.24 & OM & 36.08 & 41.53 & 39.22 & OM & 35.32 & 32.77 & 40.93 & \textbf{42.85} & \textbf{44.32} \\
\midrule
\multirow{2}*{Cornell}    & Unknown = 1  & 42.13 & 42.49 & 45.43 & 44.26 & 44.43 & 49.73 & 55.70 & 51.01 & 55.44 & 51.01 & 60.95 & 60.49 & 62.87 & \textbf{64.43} & \textbf{63.09} \\
                             & Unknown = 2 & 62.46 & 64.33 & 67.22 & 70.54 & 75.44 & 70.32 & 83.16 & 84.05 & 75.82 & 74.05 & \textbf{86.08} & 84.43 & 80.40 & 80.38 & \textbf{86.71}\\
\midrule
\multirow{2}*{Film}    & Unknown = 1  & 15.32 & 15.32 & 13.29 & 14.38 & 28.22 & 27.48 & \textbf{30.38} & 25.85 & \textbf{30.78} & 30.15 & 27.56 & 26.89 & 28.79 & 28.49 & 28.55 \\
                             & Unknown = 2 & 23.66 & 27.53 & 32.34 & 30.22 & 44.65 & 38.81 & \textbf{52.17} & 40.87 & 30.60 & 51.81 & 49.17 & 46.43 & 43.91 & \textbf{52.51} & \textbf{52.51}\\
\midrule
\multirow{2}*{Pubmed}    & Unknown = 1  & 33.48 & 43.02 & 40.25 & 47.37 & 53.59 & 55.74 & 59.53 & 60.69 & 53.72 & 59.48 & 45.95 & 52.23 & 57.45 & \textbf{61.54} & \textbf{61.45} \\
                             & Unknown = 2 & 29.12 & 35.09 & 42.86 & 39.56 & 39.12 & 34.03 & 40.88 & 42.47 & 39.12 & \textbf{60.04} & 53.89 & 51.56 & 52.49 & \textbf{63.39} & 40.07 \\
\midrule
\multirow{2}*{Tesax}    & Unknown = 1  & 35.36 & 40.41 & 45.39 & 48.76 & 44.43 & 41.74 & 55.70 & 55.03 & 55.44 & \textbf{64.43} & 45.64 & 58.99 & 61.48 & \textbf{66.44} & 63.09\\
                             & Unknown = 2 & 46.55 & 45.38 & 60.21 & 56.38 & 47.47 & 81.01 & 53.16 & 76.58 & 75.82 & 78.48 & \textbf{81.65} & 74.43 & 80.13 & 81.01 & \textbf{82.91} \\
\midrule
\multirow{2}*{UAI}    & Unknown = 1  & 44.32 & 40.75 & 42.11 & 42.44 & 39.23 & 40.75 & 47.81 & 45.64 & 47.39 & 46.72 & 54.00 & 53.56 & \textbf{54.10} & 51.83 & \textbf{55.44}\\
                             & Unknown = 2 & 37.43 & 37.98 & 40.87 & 43.45 & 40.82 & 41.69 & 41.75 & 50.22 & 50.31 & 46.99 & 42.11 & 48.10 & \textbf{50.63} & 48.79 & \textbf{51.41}\\
\midrule
\multirow{2}*{Wisconsin}    & Unknown = 1  & 40.91 & 43.22 & 45.42 & 41.67 & 52.25 & 63.37 & 61.39 & 68.81 & 64.46 & \textbf{70.30} & 66.63 & 66.14 & 61.88 & \textbf{70.30} & \textbf{73.76} \\
                             & Unknown = 2 & 43.73 & 42.14 & 50.32 & 48.93 & 53.79 & 56.31 & 59.71 & 58.74 & 64.85 & 71.36 & \textbf{72.76} & 70.49 & 62.33 & 67.48 & \textbf{73.30} \\
\bottomrule
\end{tabular}}
\label{ACCcomparsionSingleClassification}
\end{table*}

\begin{table*}[t]
\centering
\caption{Accuracy of all compared multi-modal semi-supervised classification methods under $10\%$ ratio of training labels and different unknown classes, where the best and running-up results are highlighted in bold (mean\%). OM means out-of-memory.}
\resizebox{\textwidth}{!}{
\begin{tabular}{c|c||cc|ccc|cc|cccc|ccc||cc|cc}
\toprule
\multicolumn{2}{c||}{Datasets $\backslash$ Methods} & MLP & MAE & DUA-Net & TMC-Net & DSRL-Net & GCN & GAT & SGCN & FAGCN & AMGCN & HOGGCN & GNN-LF & GNN-HF & HGNN$^{+}$ & MSL-Net & MSGL-Net & MHSL-Net & MHSGL-Net \\
\midrule
\multirow{2}*{Caltech102} & Unknown = 1 & 11.76 & 32.87 & 30.21 & 33.99 & 34.37 & 34.15 & 30.89 & 38.77 & 34.02 & 37.99 & 35.87 & 36.17 & 39.98 & 41.05 & \textbf{48.39} & 45.56 & 42.86 & \textbf{49.06}\\
                             & Unknown = 3 & 12.74 & 17.66 & 20.53 & 21.76 & 35.33 & 33.80 & 20.54 & 28.65 & 31.41 & 39.21 & 34.37 & 34.86 & 40.07 & 41.75 & \textbf{48.44} & 43.62 & 39.31 & \textbf{47.84}\\
                             \midrule
\multirow{2}*{Hdigit}    & Unknown = 1 & 45.72 & 60.96 & 73.37 & 78.65 & 88.06 & 87.08 & 90.22 & 88.44 & 88.82 & 92.75 & 85.91 & 88.34 & \textbf{92.97} & 92.15 & \textbf{93.79} & 91.98 & 92.62 & 90.81\\
                             & Unknown = 3 & 65.59 & 69.58 & 69.95 & 79.43 & 80.28 & 90.78 & 88.49 & 93.57 & 83.50 & 94.05 & 90.20 & 93.08 & 89.28 & 93.51 & \textbf{96.56} & \textbf{95.02} & 90.90 & 92.58\\
\midrule
\multirow{2}*{MITIndoor}       & Unknown = 1 & 20.70 & 19.55 & 21.47 & 22.65 & 29.41 & 17.65 & 21.33 & 32.43 & 40.56 & 37.24 & 32.91 & 37.57 & 32.42 & \textbf{42.31} & 40.26 & \textbf{42.54} & 40.70 & 39.53 \\
                             & Unknown = 3 & 22.39 & 26.67 & 26.31 & 21.47 & 31.55 & 17.82 & 24.55 & 31.26 & 41.63 & \textbf{43.70} & 40.78 & 42.70 & 40.60 & \textbf{43.97} & 39.94 & 40.26 & 42.10 & 37.60\\
\midrule
\multirow{2}*{MNIST}       & Unknown = 1  & 73.14 & 73.57 & 73.87 & 86.05 & 85.40 & 82.10 & 84.23 & 82.78 & 82.30 & 85.07 & 86.71 & 88.53 & \textbf{89.90} & 86.13 & \textbf{90.21} & 89.74 & 85.37 & 88.88 \\
                             & Unknown = 3 & 78.13 & 72.52 & 73.46 & 83.53 & 85.88 & 85.44 & 87.23 & 89.84 & 84.62 & 90.87 & 82.86 & 83.04 & 89.80 & 90.86 & \textbf{91.85} & \textbf{91.83} & 88.59 & 91.40\\
\midrule
\multirow{2}*{NoisyMNIST}       & Unknown = 1 & 50.89 & 52.43 & 55.87 & 56.41 & 72.22 & 63.64 & OM & 71.13 & 71.03 & \textbf{76.83} & OM & 68.99 & 72.56 & 74.37 & \textbf{77.89} & 71.51 & 74.38 & 67.78\\
                             & Unknown = 3 & 59.25 & 63.58 & 62.13 & 62.44 & 64.21 & 64.04 & OM & 71.52 & 75.44 & 80.71 & OM & 79.65 & 73.92 & 78.22 & \textbf{84.72} & \textbf{83.60} & 73.22 & 75.23 \\
\midrule
\multirow{2}*{NUS-WIDE}     & Unknown = 1  & 28.68 & 30.76 & 35.40 & 31.45 & 35.22 & 28.94 & 26.68 & 27.66 & 26.60 & 36.70 & 38.11 & 28.89 & 37.50 & 33.90 & \textbf{38.40} & 37.09 & \textbf{40.48} & 38.09 \\
                             & Unknown = 3 & 40.36 & 39.83 & 44.53 & 42.52 & 39.18 & 40.89 & 43.46 & 40.36 & 43.98 & 41.41 & 39.23 & 45.48 & 46.68 & 44.03 & \textbf{48.39} & 47.04 & \textbf{48.39} & 46.81 \\
\midrule
\multirow{2}*{Scene15}       & Unknown = 1  & 16.15 & 17.17 & 23.66 & 40.02 & 45.90 & 39.94 & 43.02 & 48.99 & 55.06 & 56.98 & 51.06 & 53.20 & 58.93 & 57.61 & \textbf{59.09} & 55.99 & \textbf{60.69} & 58.20 \\
                             & Unknown = 3 & 28.06 & 31.33 & 44.57 & 49.21 & 54.70 & 45.29 & 55.96 & 61.22 & 56.75 & 59.20 & 50.53 & 51.79 & 59.36 & 60.31 & \textbf{64.47} & 62.31 & \textbf{63.92} & 62.64 \\
\midrule
\multirow{2}*{Youtube}    & Unknown = 1  & 26.78 & 22.15 & 25.32 & 31.99 & 30.21 & 40.70 & 41.80 & 48.31 & 48.51 & 44.82 & 43.54 & 44.22 & 43.85 & 49.59 & \textbf{51.82} & 47.79 & \textbf{50.83} & 46.88\\
                             & Unknown = 3 & 39.02 & 38.87 & 37.78 & 43.32 & 50.01 & 40.81 & 42.85 & 41.94 & 48.90 & 45.74 & 53.54 & 49.33 & 47.71 & 58.84 & \textbf{61.40} & \textbf{61.74} & 58.02 & 56.87\\
\bottomrule
\end{tabular}}
\label{ACCcomparsionMultiClassification}
\end{table*}

\textbf{Open-world Semi-supervised Node Classification.} Table \ref{ACCcomparsionSingleClassification} records the semi-supervised node classification accuracy results under $10\%$ ratio of labeled nodes with different unknown classes, and we have obtained the following observations:
\begin{itemize}
\item The derived methods from the proposed single-modal trustworthy frameworks have achieved superior results on most datasets, including baselines and their variants.
\item These methods show good performance compared to four non-GCN-based approaches among these datasets.
This indicates that the proposed protocol exported methods can effectively improve the expression ability from feature-level.
\item The proposed frameworks demonstrate superior performance compared to four GCN and three GNN variants across most datasets.
    This can be attributed to the fact that we not only consider interpretable graph-topological dissemination structures, but also incorporate demand-driven representation learning, enabling the capture of crucial node features.
\item Overall, the well-designed trustworthy frameworks have enabled the examples to outperform competitors on open-world tasks while also ensured trustworthy results.
    Due to the transparent network design and the broad task-interface in the framework protocols, which also exhibit some ability to perceive unknown entities.
\end{itemize}

\begin{figure}[t]
  \centering
  \includegraphics[width=0.48\textwidth]{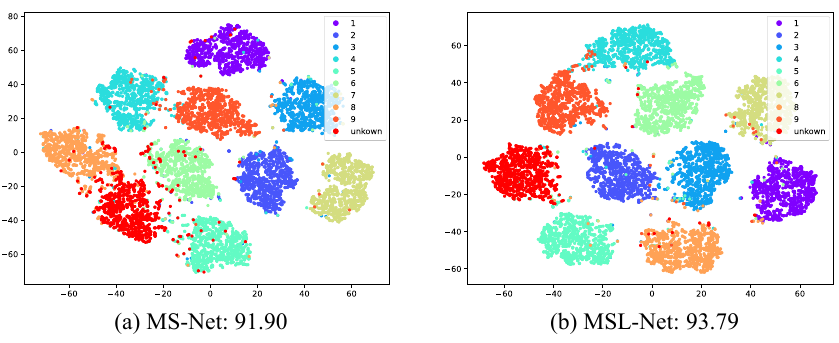}\\
  \caption{A t-SNE visualization result of MS-Net and MSL-Net on the Hdigit of the multi-modal semi-supervised classification tasks under one unknown class (Evaluation: accuracy).}
  \label{TSNEVisualizationMSLNet}
\end{figure}

\begin{figure*}[t]
  \centering
  \includegraphics[width=\linewidth]{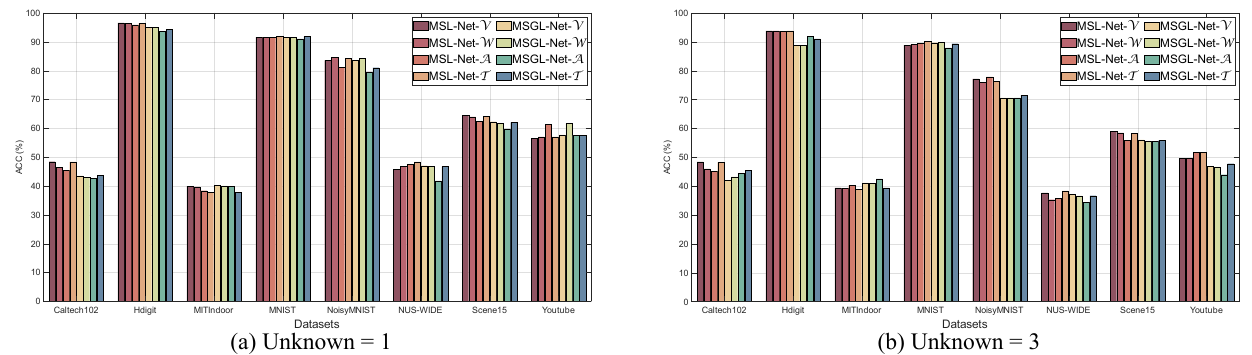}\\
  \caption{Ablation study of generalized fusion ways of MSL-Net and MSGL-Net (one and three unknown classes).}
  \label{paramultifusion1}
\end{figure*}

\textbf{Open-world Multi-modal Semi-supervised Classification.} Table \ref{ACCcomparsionMultiClassification} displays the multi-modal semi-supervised classification accuracy results under $10\%$ ratio of labeled samples with different unknown classes, and we found results in the following aspects:
\begin{itemize}
\item Following the trend of single-modal tasks, the proposed methods can still achieve good performance on multi-modal tasks compared to other multi-modal extension methods.
\item Compared with non-GNN-based, GCN, and GNN-based multi-modal methods, the proposed networks in frameworks can achieve good performance and are compatible with complex data processing capabilities.
The expressive power of multi-modal models has also been indirectly improved through the designed components.
\item The results indicate that the proposed methods can maintain trustworthiness while achieving excellent performance in complex development open-world scenarios thanks to the designed trustworthy learning frameworks.
\end{itemize}

\subsection{Trustworthiness Study}\label{sec:abla}
\begin{itemize}
\item \textbf{Interpretability of the Proposed Frameworks:} Tables 6-7 of the \textbf{Appendix} indicate that adding graph-topologial terms (including hypergraphs) to the trustworthy frameworks leads to better performance in most single-modal and multi-modal learning task cases.
    In addition, we visually display some instances in Figure \ref{TSNEVisualizationMSLNet} (for more detailed information, refer to Figures 1-2 of the \textbf{Appendix}), which present that the graph-topological frameworks perform better in recognizing unknown samples.
    These figures provide an intuitive illustration of the effectiveness of considering the graph topology structure in improving model performance with stronger recognition ability for unknown instances, thereby demonstrating the post-hoc-level interpretability of the proposed frameworks.
\item \textbf{Generalization of the Proposed Frameworks:} The previous content has already presented the generalization of the proposed frameworks in designing networks, regularizers and downstream tasks.
    Next, we will reveal the generalization of fusion in trustworthy multi-modal frameworks.
Figure \ref{paramultifusion1} proclaims that trusted fusion can achieve favorable fusion results in most situations through dynamically alleviating the impact of uncertainty caused by data heterogeneity, thus achieving positive results after fusion procedures (for more detailed information, refer to Figures 3-4 of the \textbf{Appendix}).
This is an inspiring concept: When constructing multi-modal trusted networks, we could utilize such fusion methods to promote the trustworthiness of the multi-modal open-world co-latent representation.

\item \textbf{Robustness of the Proposed Frameworks:} In addition to design robustness to open environments, model robustness is also presented here.
    The influence of hyper-parameters $\lambda_{1}$ and $\lambda_{2}$ of open-world training losses can be observed in Figures 5-8 of the \textbf{Appendix}.
    These figures depict the impact of tuning these parameters within the ranges of $[0.001, 0.01, \cdots, 100]$ on frameworks.
    The results present that the proposed frameworks demonstrate stable performance across most values, indicating their robustness.
    However, when $\lambda_{2}$ is set to a small value, it significantly affects the performance, indicating the crucial role played by the unknown loss in the generalization of the entire model.
\end{itemize}

%\begin{figure}[!htbp]
%  \centering
%  \includegraphics[width=\linewidth]{./Figures/unseen1_multi_all.pdf}\\
%  \caption{Ablation study of generalized fusion ways of MSL-Net and MSGL-Net.}
%  \label{paramultifusion1}
%\end{figure}

\subsection{Parameter Sensitivity}\label{sec:para}

\begin{figure}[!htbp]
  \centering
  \includegraphics[width=0.48\textwidth]{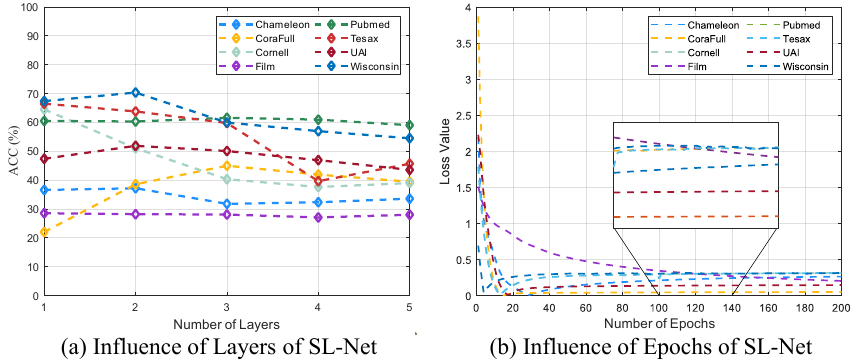}\\
  \caption{Parameter sensitivity analysis of SL-Net on the layer impact and convergence behavior.}
  \label{blocklossclassification}
\end{figure}

\begin{itemize}
\item \textbf{Impact of Layers:} Figure \ref{blocklossclassification} (a) depicts the impact of the number of layers on performance (for more detailed information, refer to Figure 9 of the \textbf{Appendix}).
    Generally, as the number of layers increases, performance stabilizes after less than four layers, and in some situations, performance decreases accordingly.
    This observation suggests that the proposed frameworks exhibit some degree of robustness.

\item \textbf{Convergence Analysis:}  Figure \ref{blocklossclassification} (b) reveals that loss values plummet and eventually converge across all test datasets (for more detailed information, refer to Figure 10 of the \textbf{Appendix}).
    Moreover, ACC tends to be steady and fluctuates gradually as the loss value converges in the single-modal and multi-modal semi-supervised classification tasks.
    All of these phenomena attest to the convergence behaviors of the proposed frameworks.
\end{itemize}

%%%%%%%%%%%%%%%%%%%%%%%%%%%%%%%%%%%%%%%%%%%%%%%%%%%%%%%%%%%%%%%%%%%%%%%%%%%%%%%%%%%%%%%%%%%%%%%%%%%%%%

\section{Conclusion and Future Work}\label{Conclusion}

In this paper, we proposed a novel perspective on providing a family of design neural approaches for bridging trustworthiness and open-world learning exploration, which included the accomplishment of enhancing various trustworthy properties.
We explosively enhanced trustworthiness, including interpretability, generalization, and robustness by the establishment of design-level explainability, environmental well-being task-interfaces and open-world recognition programs.
By following the proposed trustworthy open-world protocols, we could develop methods that performed well across a wide range of applications while maintaining trustworthiness.
Extensive experiments across fields involved in each trustworthy property have demonstrate that our exploration for the open-world trustworthy frameworks presented in this paper could provide readers with more valuable insights.
In future, we will consider how to bridge trusted and open-world under unsupervised scenarios.

\clearpage

\title{Bridging Trustworthiness and Open-World Learning: \\An Exploratory Neural Approach for Enhancing \\ Interpretability, Generalization, and Robustness (Appendix)}

\setcounter{section}{0}

\section{Related Work on Trustworthy Learning}\label{sec:rel}

Hence, we have presented the relevance of trustworthiness in AI.
Immediately, we will comprehensively review the exploration on serval properties of trustworthy learning.

\textbf{Definition of Trustworthy Learning:} Trustworthy learning attempt to construct a framework that can guarantee both task performance and human-environment-friendly trustworthiness \cite{Li2023Trustworthy}.

\textbf{Interpretability:} Interpretability of trustworthy learning means that the systems should be transparent for human, that is, it is necessary for human that we need to know the process how a model learns patterns and makes decisions \cite{Arrieta2020Explainable}.
All in all, the concerned research classify this conception into two levels: design-level \cite{Zhang2022Protgnn, Xie2022Optimization, Li22Optimization} (this paper also tries to pursue) and post-hoc-level \cite{Ribeiro18Marco, Abrate21Counterfactual, Lin22OrphicX} interpretability.

\textbf{Generalization:} Generalization of trustworthy learning implies that the model general and cognitive capabilities covering more situations \cite{Zhang2020Generalized, Li2022OOD-GNN, Zhou2023Domain}, including limited data, complex environments, domain shift and etc.

\textbf{Robustness:} Robustness of trustworthy learning refers that the ability of the systems to handle open-world environments or various unknown data instances \cite{Sun20Adversarial, Boult21Towards, Li2023Trustworthy}.

\textbf{Limitations, Solutions and Beyond:} Although they have done loads of excellent research on various properties of trustworthy learning, there are still few attempts to consider these properties into a universal framework.
It is more laborious and challenging to explore effective architectures that could perform better and generalize well to trustworthy learning.
To this end, we bent over backwards to accomplish this by presenting a comprehensive trustworthy framework protocol capable of embracing more trustworthiness properties in an executive setting.
Beyond that, we also promote the integration of trustworthy learning framework from a universal perspective, which is exactly the main contributions of this work.
Given the extensive research fields involved in each trustworthy property, we aim to provide readers with some ideas in this regard through the superficial exploration of the trustworthy frameworks presented in this paper.

\section{Framework Supplementary}

In this section, we provide some supplementary details about the full name, derivation and other details of the example networks implied by the proposed frameworks.

\subsection{Full Name of the Proposed Abbreviated Network}
Here, we give the full name of the proposed abbreviated networks of this paper in Tables 1-3.
\begin{itemize}
\item S-Net: Sparse Network.
\item SG-Net: Sparse Group Network.
\item SL-Net: Sparse Learning Network.
\item SGL-Net: Sparse Group Learning Network.
\item MS-Net: Multi-modal Sparse Network.
\item MSG-Net: Multi-modal Sparse Group Network.
\item MSL-Net: Multi-modal Sparse Learning Network.
\item MHSL-Net: Multi-modal Hypergraph Sparse Learning Network.
\item MSGL-Net: Multi-modal Sparse Group Learning Network.
\item MHSGL-Net: Multi-modal Hypergraph Sparse Group Learning Network.
\end{itemize}

To provide a unified framework for both single-modal and multi-modal learning, the proposed method can incorporate various networks such as S-Net, SG-Net, SL-Net, and SGL-Net, which focus on learning single-modal open-world latent representations using features and graphs for downstream tasks. 
Additionally, the proposed method can also incorporate networks such as MS-Net, MSG-Net, MSL-Net, MHSL-Net, MSGL-Net, and MHSGL-Net, which are designed for learning co-latent representations based on multi-modal features and graphs for downstream tasks, where the adjacency graph of each modal is constructed by $k$-NN methods.

\subsection{Framework Examples and Derivations}

Here, we reveal several examples for better understanding and instantiating generalized Framework (6)-(7) to (15)-(16).

\textbf{SL-Net.} First, instantiating optimization-inspired objective function under Problem (2) of SL-Net is
\begin{equation}\label{SolvingZExample1}
\min\limits_{\mathbf{Z}}\frac{1}{2}\Big(\|\mathbf{X}-\mathbf{Z}\mathbf{D}\|_{F}^{2}+\alpha\operatorname{Tr}(\mathbf{Z}^{T} \mathbf{L} \mathbf{Z})\Big)+\beta\|\mathbf{Z}\|_{1}.
\end{equation}
Subsequently, we utilize machine optimization equation (5) in the main paper to solve the above problem as
\begin{equation}\label{SolvingZExample2}
\begin{aligned}
{\mathbf{Z}}^{(t+1)}&= \operatorname{\textbf{Prox}}_{\frac{\beta}{L}}\Big(\mathbf{Z}^{(t)}-\frac{1}{{L}}(-\mathbf{X}\mathbf{D}^{T}+\mathbf{Z}^{(t)}\mathbf{D}^{T}\mathbf{D}
+\alpha\mathbf{L}\mathbf{Z}^{(t)})\Big)\\
&=\operatorname{\textbf{Prox}}_{\frac{\beta}{L}}\Big(\mathbf{Z}^{(t)}( \mathbf{I}-\frac{1}{L}\mathbf{D}^{T}\mathbf{D}) -\frac{\alpha}{L}\mathbf{L}\mathbf{Z}^{( t)}+\frac{1}{L}\mathbf{X}\mathbf{D}^{T} \Big).
\end{aligned}
\end{equation}
Here, we link the above updating rules to construct single-modal networks under Framework (7) as
\begin{equation}\label{SolvingZExample3}
{\mathbf{Z}}^{(t+1)} \leftarrow \mathcal{H}_{\theta}\left(\mathbf{Z}^{\left( t\right)}\mathbf{F}- \alpha\mathbf{L}\mathbf{Z}^{\left(t \right)}\mathbf{W}+\mathbf{X}\mathbf{U} \right),
\end{equation}
where $\mathbf{F}=\mathbf{I}-\frac{1}{L}\mathbf{D}^{T}\mathbf{D}$, $\mathbf{W}=\frac{1}{L}\mathbf{I}$, $\mathbf{U}=\frac{1}{L}\mathbf{D}^{T}$, $\theta = \frac{\beta}{L}$ and $\mathcal{H}_{\theta}=\operatorname{\textbf{Prox}}_{\frac{\beta}{L}}$, and $\mathbf{I}$ is an identity matrix.
Particularly, for these sparsity and row sparsity parameterized regularizers $\mathcal{H}_{\theta}=\operatorname{\textbf{Prox}}_{\frac{\beta}{L}}$ or $\mathcal{S}_{\theta}=\operatorname{\textbf{Prog}}_{\frac{\beta}{L}}$ of $\mathcal{P}_{\theta}$ instantiation in Table 1 of the main body, we have
\begin{equation}\label{multiviewsparsesolve}
\mathcal{H}_{\theta}=F(z_i-\theta)-F(-z_i-\theta), \text{and}
\end{equation}

\begin{equation}\label{multivieww21solve}%W^*(:, i)=
\mathcal{S}_{\theta}= \begin{cases}\frac{F(\left\|z_i\right\|_F-\theta)}{\left\|z_i\right\|_F} z_i, & \text { if } \theta<\left\|z_i\right\|_F, \\ 0, & \text { otherwise, }\end{cases}
\end{equation}
where $z_i$ is the $i$-th column of $\mathbf{Z} = [z_1; z_2; \cdots; z_i; \cdots]$, and $F$ can be activation functions such as ReLU, SeLU, ELU, and etc.
The above equation is an example form of the proposed single-modal generalized Framework (7).
Other networks derived from Frameworks (6) or (7) in Table 1 can also be derived based on this, and Protocol 1 can be executed for trustworthy learning.

\textbf{MSGL-Net.} Then, instantiating optimization-inspired objective function under Problem (2) of MSGL-Net is
\begin{equation}\label{SolvingZExample11}
\min\limits_{\mathbf{Z}_{m}}\sum\limits_{{m=1}}^{M}\frac{1}{2}\Big(\|\mathbf{X}_{m}-\mathbf{Z}_{m}\mathbf{D}_{m}\|_{F}^{2}+\alpha\operatorname{Tr}(\mathbf{Z}_{m}^{T} \mathbf{L}_{m} \mathbf{Z}_{m})\Big)+\beta\|\mathbf{Z}_{m}\|_{2, 1}.
\end{equation}
Similarly, machine optimization equation (5) is also used to tackle the above problem as
\begin{equation}\label{SolvingZExample22}
\begin{aligned}
{\mathbf{Z}}_{m}^{(t+1)}&= \operatorname{\textbf{Prog}}_{\frac{\beta}{L}}\Big(\mathbf{Z}_{m}^{(t)}-\frac{1}{{L}}(-\mathbf{S}_{m}\mathbf{D}_{m}^{T}+\mathbf{Z}^{(t)}\mathbf{D}_{m}^{T}\mathbf{D}_{m}+\alpha\mathbf{L}_{m}\mathbf{Z}_{m}^{(t)})\Big)\\
&=\operatorname{\textbf{Prog}}_{\frac{\beta}{L}}\Big(\mathbf{Z}_{m}^{(t)}( \mathbf{I}-\frac{1}{L}\mathbf{D}_{m}^{T}\mathbf{D}_m) -\frac{\alpha}{L}\mathbf{L}_m\mathbf{Z}_{m}^{( t)}+\frac{1}{L}\mathbf{S}_m\mathbf{D}_{m}^{T} \Big).
\end{aligned}
\end{equation}
Then, the row sparsity based sub-variate heterogeneous representation network with partial parameterization can be ensured as
\begin{equation}\label{SolvingZExample33}
{\mathbf{Z}}_{m}^{(t+1)} \leftarrow \mathcal{S}_{\theta}\left(\mathbf{Z}_{m}^{\left( t\right)}\mathbf{F}- \alpha\mathbf{L}_m\mathbf{Z}_{m}^{\left(t \right)}\mathbf{W}+\mathbf{S}_m\mathbf{U} \right),
\end{equation}
where $\mathbf{F}=\mathbf{I}-\frac{1}{L}\mathbf{D}_{m}^{T}\mathbf{D}_{m}$, $\mathbf{W}=\frac{1}{L}\mathbf{I}$, $\mathbf{U}=\frac{1}{L}\mathbf{D}_{m}^{T}$, $\theta = \frac{\beta}{L}$ and $\mathcal{S}_{\theta}=\operatorname{\textbf{Prog}}_{\frac{\beta}{L}}$, and $\mathbf{I}$ is an identity matrix.
When we obtain the sub-modal latent representation, we utilize a generalized fusion to obtain the co-latent representation $\mathbf{Z}$ as
\begin{equation}\label{SolvingZFusion}
{\mathbf{Z}}^{(t+1)} \leftarrow \mathcal{F}({\mathbf{Z}}_{1}^{(t+1)}, \cdots, {\mathbf{Z}}_{m}^{(t+1)}; \mathbf{\Theta}_{\mathcal{F}}).
\end{equation}
This finishes a complete derivation example of MSL-Net, which learns a co-latent representation.
The above equation is also an example form of the proposed multi-modal generalized Framework (16).
Other networks derived from Frameworks (15) or (16) in Table 3 can also be derived based on this, and Protocol 2 can be executed for trustworthy learning.
The remaining instance single-modal (S-Net, SG-Net, and SGL-Net) and multi-modal (MS-Net, MSG-Net, MHSL-Net, MSGL-Net, and MHSGL-Net) networks implied in Frameworks (6)-(7) to (15)-(16) have similar derivation ideas.

\subsection{Supplementary of Open-world Losses}\label{sec:tow}

It should be noted that we do not minimize loss $\mathcal{L}_{u}$, as our objective is to attain recognition by maximizing the uncertainty of the unknown classes.
%Furthermore, we rank the normalized open-world probability representations and discard samples whose ranking values fall within the bottom and top 10$\%$, respectively.
%This is because large values are favorable for recognition, while small values indicate a balanced output across each visible class, which is more conducive to enhancing the model robustness.
Lastly, we utilize the remaining samples to maximize the loss.
Through training with loss function (10), we aim to increase the discriminatory power of the recognized classes by labeling the data, while simultaneously maximizing the uncertainty loss to achieve a more balanced output for each sample, which assists in detecting unknown classes.

\subsection{Supplementary of Agent Selection}\label{sec:agent}

The question that arises is how to determine the agent.
To address this, we utilize the validation set for agent selection.
%We provide the assumption that unknown classes do not appear in the training set.
Similarly, for the samples in the validation set, we perform the same process as in training.
So we calculate the largest probability of all samples and average them to obtain $a_k$.
We still select the top 10$\%$ samples with the highest entropy as the expected unknown class samples, and their average probability is denoted by $a_u$.
Finally, we obtain the final agent value by averaging the two, i.e., $a = (a_k+a_u)/2$.
Up to the present, the open-world losses and agents are employed to perceive unknown data to assist improve the robustness of the overall framework.

\textbf{\setcounter{table}{0}}
\begin{table*}[!htbp]
\centering
\caption{Demand-driven and graph-topological layers and their generalized forms derived from the proposed frameworks.}
\label{NetworkExamples}
\resizebox{0.9\textwidth}{!}{
\begin{tabular}{c|c|c|c|c}
\toprule
  \multicolumn{1}{c|}{Abbreviations} & {Theoretical Support} & {Optimization Problems} & {Iteration Rules} & {Inspired Networks}\\
\midrule
  \multirow{1}*{S-Net}    & Beck \textit{et al.} \cite{BeckTeboulle09AFast} &   $\min\limits_{\mathbf{Z}}\frac{1}{2}\|\mathbf{X}-\mathbf{Z}\mathbf{D}\|_{F}^{2}+\alpha\|\mathbf{Z}\|_{1}$&
  $\operatorname{\textbf{Prox}}_{\frac{\alpha}{L}}\left(\mathbf{Z}^{(t)}(\mathbf{I} - \frac{1}{L}\mathbf{D}^{T}\mathbf{D}) + \frac{1}{L}\mathbf{X}\mathbf{D}^{T}\right)$ & $\mathcal{H}_{\theta}\left(\mathbf{Z}^{(t)}\mathbf{F} + \mathbf{X}\mathbf{U}\right)$\\
  \multirow{1}*{SG-Net}  & Liu \textit{et al.} \cite{Liu10Robust}  &   $\min\limits_{\mathbf{Z}}\frac{1}{2}\|\mathbf{X}-\mathbf{Z}\mathbf{D}\|_{F}^{2}+\alpha\|\mathbf{Z}\|_{2, 1}$ & $\operatorname{\textbf{Prog}}_{\frac{\alpha}{L}}\left(\mathbf{Z}^{(t)}(\mathbf{I} - \frac{1}{L}\mathbf{D}^{T}\mathbf{D}) + \frac{1}{L}\mathbf{X}\mathbf{D}^{T}\right)$ & $\mathcal{S}_{\theta}\left(\mathbf{Z}^{(t)}\mathbf{F} + \mathbf{X}\mathbf{U}\right)$\\
  \midrule
  \multirow{1}*{Our}  & Above  &  $M_g(\mathbf{x}):= M_f(\mathbf{x})+g(\mathbf{z})$ & Rule (5) & Framework (6) with example (13)\\
  %\multirow{1}*{HSL-Net \cite{Feng19Hypergraph}}     &   $\arg\min\limits_{\mathbf{Z}}\frac{1}{2}\Big(\|\mathbf{X}-\mathbf{Z}\mathbf{D}\|_{F}^{2}+\alpha\operatorname{Tr}(\mathbf{Z}^{T} \mathbf{\tilde{L}} \mathbf{Z})\Big)+\beta\|\mathbf{Z}\|_{1}$ & $\operatorname{\textbf{Prox}}_{\frac{\beta}{L}}\left( \mathbf{Z}^{\left(t\right)}\left( \mathbf{I}-\frac{1}{L}\mathbf{D}^{T}\mathbf{D} \right) -\frac{\alpha}{L}\mathbf{\tilde{L}}\mathbf{Z}^{\left(t\right)}+\frac{1}{L}\mathbf{X}\mathbf{D}^{T} \right)$ & $\mathcal{H}_{\theta}\left(\mathbf{Z}^{\left( t \right)}\mathbf{F}- \mathbf{\tilde{L}}\mathbf{Z}^{\left(t \right)}\mathbf{W}+\mathbf{X}\mathbf{U} \right)$\\
  \midrule
    \multirow{1}*{SL-Net}  & Fang \textit{et al.} \cite{Fang2023DBONet}  &   $\min\limits_{\mathbf{Z}}\frac{1}{2}\Big(\|\mathbf{X}-\mathbf{Z}\mathbf{D}\|_{F}^{2}+\alpha\operatorname{Tr}(\mathbf{Z}^{T} \mathbf{L} \mathbf{Z})\Big)+\beta\|\mathbf{Z}\|_{1}$ & $\operatorname{\textbf{Prox}}_{\frac{\beta}{L}}\left( \mathbf{Z}^{\left(t\right)}\left( \mathbf{I}-\frac{1}{L}\mathbf{D}^{T}\mathbf{D} \right) -\frac{\alpha}{L}\mathbf{L}\mathbf{Z}^{\left(t\right)}+\frac{1}{L}\mathbf{X}\mathbf{D}^{T} \right)$ & $\mathcal{H}_{\theta}\left(\mathbf{Z}^{\left( t \right)}\mathbf{F}- \alpha\mathbf{L}\mathbf{Z}^{\left(t \right)}\mathbf{W}+\mathbf{X}\mathbf{U} \right)$\\
  \multirow{1}*{SGL-Net} & Zhuang \textit{et al.} \cite{Zhuang2012Non}   &   $\min\limits_{\mathbf{Z}}\frac{1}{2}\Big(\|\mathbf{X}-\mathbf{Z}\mathbf{D}\|_{F}^{2}+\alpha\operatorname{Tr}(\mathbf{Z}^{T} \mathbf{L} \mathbf{Z})\Big)+\beta\|\mathbf{Z}\|_{2, 1}$ & $\operatorname{\textbf{Prog}}_{\frac{\beta}{L}}\left( \mathbf{Z}^{\left( t \right)}\left( \mathbf{I}-\frac{1}{L}\mathbf{D}^{T}\mathbf{D} \right) -\frac{\alpha}{L}\mathbf{L}\mathbf{Z}^{\left( t \right)}+\frac{1}{L}\mathbf{X}\mathbf{D}^{T} \right)$ & $\mathcal{S}_{\theta}\left(\mathbf{Z}^{\left( t \right)}\mathbf{F}- \alpha\mathbf{L}\mathbf{Z}^{\left(t \right)}\mathbf{W}+\mathbf{X}\mathbf{U} \right)$\\
  %\multirow{1}*{HSGL-Net \cite{Feng19Hypergraph}}     &   $\arg\min\limits_{\mathbf{Z}}\frac{1}{2}\Big(\|\mathbf{X}-\mathbf{Z}\mathbf{D}\|_{F}^{2}+\alpha\operatorname{Tr}(\mathbf{Z}^{T} \mathbf{\tilde{L}} \mathbf{Z})\Big)+\beta\|\mathbf{Z}\|_{2, 1}$ & $\operatorname{\textbf{Prog}}_{\frac{\beta}{L}}\left( \mathbf{Z}^{\left( t \right)}\left( \mathbf{I}-\frac{1}{L}\mathbf{D}^{T}\mathbf{D} \right) -\frac{\alpha}{\tilde{L}}\mathbf{\tilde{L}}\mathbf{Z}^{\left( t \right)}+\frac{1}{L}\mathbf{X}\mathbf{D}^{T} \right)$ & $\mathcal{S}_{\theta}\left(\mathbf{Z}^{\left( t \right)}\mathbf{F}- \mathbf{\tilde{L}}\mathbf{Z}^{\left(t \right)}\mathbf{W}+\mathbf{X}\mathbf{U} \right)$\\
\midrule
  \multirow{1}*{Our}  & Above  &  $M_h(\mathbf{x}):= M_f(\mathbf{x})+g(\mathbf{z})+h(\mathbf{z})$ & Rule (5) & Framework (7) with example (14)\\
\bottomrule
\end{tabular}}
\end{table*}

\begin{table*}[!htbp]
\centering
\caption{Graph-topological regularizer network layers and their generalized forms derived from the proposed frameworks, where $\textbf{P}$ is a proximal operation, and $B(\cdot)$ and $N(\cdot)$ are generalized terms about representations and features, respectively.}
\label{GraphNetworkExamples2}
\resizebox{0.9\textwidth}{!}{
\begin{tabular}{c|c|c|c|c}
\toprule
  \multicolumn{1}{c|}{Abbreviations} & {Theoretical Support} & {Optimization Problems} & {Iteration Rules} & {Inspired Networks}\\
\midrule
  \multirow{1}*{GCN/SGC}  & Kipf \textit{et al.} \cite{Kipf17Semi}  &   $\min\limits_{\mathbf{Z}}\frac{1}{2}\Big(f(\mathbf{Z})+\alpha\operatorname{Tr}(\mathbf{Z}^{T} \mathbf{L} \mathbf{Z})\Big)+\beta g(\mathbf{Z})$ & $\operatorname{\textbf{P}}_{\frac{\beta}{L}}\left( B(\mathbf{Z}^{\left( t \right)}) -\frac{\alpha}{L}\mathbf{L}\mathbf{Z}^{\left(t\right)}+N(\mathbf{X}) \right)$ & $\mathcal{P}_{\theta}\left(\mathbf{Z}^{\left( t \right)}\mathbf{F}- \alpha\mathbf{L}\mathbf{Z}^{\left(t \right)}\mathbf{W}+\mathbf{X}\mathbf{U} \right)$\\
  \multirow{1}*{JKNet}  & Klicpera \textit{et al.} \cite{Klicpera19Predict}  &   $\min\limits_{\mathbf{Z}}\frac{1}{2}\Big(f(\mathbf{Z})+\alpha\operatorname{Tr}\left(\mathbf{Z}^{T}\sum_{k=1}^K \gamma_k \mathbf{A}^k \mathbf{Z}\right)\Big)+\beta g(\mathbf{Z})$ & $\operatorname{\textbf{P}}_{\frac{\beta}{L}}\left( B(\mathbf{Z}^{\left( t \right)})  -\frac{\alpha}{L}\sum_{k=1}^K \gamma_k \mathbf{A}^k \mathbf{Z}^{\left( t \right)}+N(\mathbf{X}) \right)$ & $\mathcal{P}_{\theta}\left(\mathbf{Z}^{\left( t \right)}\mathbf{F}- \alpha\sum_{k=1}^K \gamma_k \mathbf{A}^k \mathbf{Z}^{\left(t \right)}\mathbf{W}+\mathbf{X}\mathbf{U} \right)$\\
  \multirow{1}*{GNN-LF} & Zhu \textit{et al.} \cite{Zhu2021interpreting}   &   $\min\limits_{\mathbf{Z}}\frac{1}{2}\Big(f(\mathbf{Z})+\alpha\operatorname{Tr}(\mathbf{Z}^{\top}(\mathbf{I}+\delta \mathbf{A})^{-1}(\mathbf{I}+\gamma \mathbf{A}) \mathbf{Z})\Big)+\beta g(\mathbf{Z})$ & $\operatorname{\textbf{P}}_{\frac{\beta}{L}}\left( B(\mathbf{Z}^{\left( t \right)})  -\frac{\alpha}{L}(\mathbf{I}+\delta \mathbf{A})^{-1}(\mathbf{I}+\gamma \mathbf{A})\mathbf{Z}^{\left( t \right)}+N(\mathbf{X}) \right)$ & $\mathcal{P}_{\theta}\left(\mathbf{Z}^{\left( t \right)}\mathbf{F}- \alpha(\mathbf{I}+\delta \mathbf{A})^{-1}(\mathbf{I}+\gamma \mathbf{A})\mathbf{Z}^{\left(t \right)}\mathbf{W}+\mathbf{X}\mathbf{U} \right)$\\
  \multirow{1}*{GNN-HF} & Zhu \textit{et al.} \cite{Zhu2021interpreting}   &   $\min\limits_{\mathbf{Z}}\frac{1}{2}\Big(f(\mathbf{Z})+\alpha\operatorname{Tr}(\mathbf{Z}^{\top}(\mathbf{I}+\delta \mathbf{L})^{-1}(\mathbf{I}+\gamma \mathbf{L}) \mathbf{Z})\Big)+\beta g(\mathbf{Z})$ & $\operatorname{\textbf{P}}_{\frac{\beta}{L}}\left( B(\mathbf{Z}^{\left( t \right)}) -\frac{\alpha}{L}(\mathbf{I}+\delta \mathbf{L})^{-1}(\mathbf{I}+\gamma \mathbf{L})\mathbf{Z}^{\left( t \right)}+N(\mathbf{X}) \right)$ & $\mathcal{P}_{\theta}\left(\mathbf{Z}^{\left( t \right)}\mathbf{F}- \alpha(\mathbf{I}+\delta \mathbf{L})^{-1}(\mathbf{I}+\gamma \mathbf{L})\mathbf{Z}^{\left(t \right)}\mathbf{W}+\mathbf{X}\mathbf{U} \right)$\\
  \multirow{1}*{MHGCN} & Yu \textit{et al.} \cite{Yu2022Multiplex}   &   $\min\limits_{\mathbf{Z}}\frac{1}{2}\Big(f(\mathbf{Z})+\alpha\operatorname{Tr}(\mathbf{Z}^{\top}\frac{1}{l} \sum_{i=1}^l \mathbf{A}\mathbf{Z})\Big)+\beta g(\mathbf{Z})$ & $\operatorname{\textbf{P}}_{\frac{\beta}{L}}\left( B(\mathbf{Z}^{\left( t \right)}) -\frac{\alpha}{L}(\frac{1}{l} \sum_{i=1}^l \mathbf{A})\mathbf{Z}^{\left( t \right)}+N(\mathbf{X}) \right)$ & $\mathcal{P}_{\theta}\left(\mathbf{Z}^{\left( t \right)}\mathbf{F}- \alpha(\frac{1}{l} \sum_{i=1}^l \mathbf{A})\mathbf{Z}^{\left(t \right)}\mathbf{W}+\mathbf{X}\mathbf{U} \right)$\\
    \multirow{1}*{HOGGCN} & Wang \textit{et al.} \cite{Wang22Powerful}   &   $\min\limits_{\mathbf{Z}}\frac{1}{2}\Big(f(\mathbf{Z})+\alpha\operatorname{Tr}(\mathbf{Z}^{\top}(\delta +\gamma \hat{\mathbf{D}}^{-1} \mathbf{A}_k \odot \mathbf{H}) \mathbf{Z})\Big)+\beta g(\mathbf{Z})$ & $\operatorname{\textbf{P}}_{\frac{\beta}{L}}\left( B(\mathbf{Z}^{\left( t \right)}) -\frac{\alpha}{L}(\delta +\gamma \hat{\mathbf{D}}^{-1} \mathbf{A}_k \odot \mathbf{H})\mathbf{Z}^{\left( t \right)}+N(\mathbf{X}) \right)$ & $\mathcal{P}_{\theta}\left(\mathbf{Z}^{\left( t \right)}\mathbf{F}- \alpha(\delta +\gamma \hat{\mathbf{D}}^{-1} \mathbf{A}_k \odot \mathbf{H})\mathbf{Z}^{\left(t \right)}\mathbf{W}+\mathbf{X}\mathbf{U} \right)$\\
\midrule
  \multirow{1}*{Our}  & Above  &  $M_h(\mathbf{x}):= M_f(\mathbf{x})+g(\mathbf{z})+h(\mathbf{z})$
 & Rule (5) & Framework (7) with example (14)\\
\bottomrule
\end{tabular}}
\end{table*}

\begin{table*}[!htbp]
\centering
\caption{Multi-modal network layers and their generalized forms derived from the proposed frameworks.}
\label{NetworkExamples3}
\resizebox{0.9\textwidth}{!}{
\begin{tabular}{c|c|c|c|c}
\toprule
  \multicolumn{1}{c|}{Abbreviations} & {Theoretical Support} & {Optimization Problems} & {Iteration Rules} & {Inspired Networks}\\
\midrule
  \multirow{1}*{MS-Net} &   Beck \textit{et al.} \cite{BeckTeboulle09AFast}   &   $\min\limits_{\mathbf{Z}_{m}}\sum\limits_{{m=1}}^{M}\frac{1}{2}\|\mathbf{X}_{m}-\mathbf{Z}_{m}\mathbf{D}_{m}\|_{F}^{2}+\alpha\|\mathbf{Z}_{m}\|_{1}$&
  $\frac{1}{M}\sum\limits_{{m=1}}^{M}\operatorname{\textbf{Prox}}_{\frac{\alpha}{L}}\left(\mathbf{Z}_{m}^{(t)}(\mathbf{I} - \frac{1}{L}\mathbf{D}_{m}^{T}\mathbf{D}_{m}) + \frac{1}{L}\mathbf{X}_{m}\mathbf{D}_{m}^{T}\right)$ & $\mathcal{T}\left(\mathcal{H}_{\theta}\left(\mathbf{Z}_{m}^{(t)}\mathbf{F} + \mathbf{X}_{m}\mathbf{U}_{m}\right)\right)$\\
    \multirow{1}*{MSG-Net}  & Liu \textit{et al.} \cite{Liu10Robust}  &   $\min\limits_{\mathbf{Z}_{m}}\sum\limits_{{m=1}}^{M}\frac{1}{2}\|\mathbf{X}_{m}-\mathbf{Z}_{m}\mathbf{D}_{m}\|_{F}^{2}+\alpha\|\mathbf{Z}_{m}\|_{2, 1}$ & $\frac{1}{M}\sum\limits_{{m=1}}^{M}\operatorname{\textbf{Prog}}_{\frac{\alpha}{L}}\left(\mathbf{Z}_{m}^{(t)}(\mathbf{I} - \frac{1}{L}\mathbf{D}_{m}^{T}\mathbf{D}_{m}) + \frac{1}{L}\mathbf{X}_{m}\mathbf{D}_{m}^{T}\right)$ & $\mathcal{T}\left(\mathcal{S}_{\theta}\left(\mathbf{Z}_{m}^{(t)}\mathbf{F} + \mathbf{X}_{m}\mathbf{U}_{m}\right)\right)$\\
  \midrule
  \multirow{1}*{Our}  & Above  &  $M_g(\mathbf{x}_{m}):= \sum\limits_{{m=1}}^{M} M_f(\mathbf{x}_{m})+g(\mathbf{z}_{m})$ & Rule (5) & Framework (15) with examples (18)-(19)\\
    \midrule
  \multirow{1}*{MSL-Net}  &   Fang \textit{et al.} \cite{Fang2023DBONet}  &   $\min\limits_{\mathbf{Z}_{m}}\sum\limits_{{m=1}}^{M}\frac{1}{2}\Big(\|\mathbf{X}_{m}-\mathbf{Z}_{m}\mathbf{D}_{m}\|_{F}^{2}+\alpha\operatorname{Tr}(\mathbf{Z}_{m}^{T} \mathbf{L}_{m} \mathbf{Z}_{m})\Big)+\beta\|\mathbf{Z}_{m}\|_{1}$ & $\frac{1}{M}\sum\limits_{{m=1}}^{M}\operatorname{\textbf{Prox}}_{\frac{\beta}{L}}\left( \mathbf{Z}_{m}^{\left(t\right)}\left( \mathbf{I}-\frac{1}{L}\mathbf{D}_{m}^{T}\mathbf{D}_m \right) -\frac{\alpha}{L}\mathbf{L}_m\mathbf{Z}_{m}^{\left(t\right)}+\frac{1}{L}\mathbf{X}_m\mathbf{D}_{m}^{T} \right)$ & $\mathcal{T}\left(\mathcal{H}_{\theta}\left(\mathbf{Z}_{m}^{\left( t \right)}\mathbf{F}- \alpha\mathbf{L}_m\mathbf{Z}_{m}^{\left(t \right)}\mathbf{W}+\mathbf{X}_m\mathbf{U}_{m} \right)\right)$\\
  \multirow{1}*{MHSL-Net} & Feng \textit{et al.} \cite{Feng19Hypergraph}   &   $\min\limits_{\mathbf{Z}_{m}}\sum\limits_{{m=1}}^{M}\frac{1}{2}\Big(\|\mathbf{X}_{m}-\mathbf{Z}_{m}\mathbf{D}_{m}\|_{F}^{2}+\alpha\operatorname{Tr}(\mathbf{Z}_{m}^{T} \mathbf{\tilde{L}}_{m} \mathbf{Z}_{m})\Big)+\beta\|\mathbf{Z}_{m}\|_{1}$ & $\frac{1}{M}\sum\limits_{{m=1}}^{M}\operatorname{\textbf{Prox}}_{\frac{\beta}{L}}\left( \mathbf{Z}_{m}^{\left(t\right)}\left( \mathbf{I}-\frac{1}{L}\mathbf{D}_{m}^{T}\mathbf{D}_m \right) -\frac{\alpha}{L}\mathbf{\tilde{L}}_m\mathbf{Z}_{m}^{\left(t\right)}+\frac{1}{L}\mathbf{X}_m\mathbf{D}_{m}^{T} \right)$ & $\mathcal{T}\left(\mathcal{H}_{\theta}\left(\mathbf{Z}_{m}^{\left( t \right)}\mathbf{F}- \alpha\mathbf{\tilde{L}}_m\mathbf{Z}_{m}^{\left(t \right)}\mathbf{W}+\mathbf{X}_m\mathbf{U}_{m} \right)\right)$\\
  \multirow{1}*{MSGL-Net}  & Zhuang \textit{et al.} \cite{Zhuang2012Non}   &   $\min\limits_{\mathbf{Z}_{m}}\sum\limits_{{m=1}}^{M}\frac{1}{2}\Big(\|\mathbf{X}_{m}-\mathbf{Z}_{m}\mathbf{D}_{m}\|_{F}^{2}+\alpha\operatorname{Tr}(\mathbf{Z}_{m}^{T} \mathbf{L}_{m} \mathbf{Z}_{m})\Big)+\beta\|\mathbf{Z}_{m}\|_{2, 1}$ & $\frac{1}{M}\sum\limits_{{m=1}}^{M}\operatorname{\textbf{Prog}}_{\frac{\beta}{L}}\left( \mathbf{Z}_{m}^{\left( t \right)}\left( \mathbf{I}-\frac{1}{L}\mathbf{D}_{m}^{T}\mathbf{D}_m \right) -\frac{\alpha}{L}\mathbf{L}_m\mathbf{Z}_{m}^{\left( t \right)}+\frac{1}{L}\mathbf{X}_m\mathbf{D}_{m}^{T} \right)$ & $\mathcal{T}\left(\mathcal{S}_{\theta}\left(\mathbf{Z}_{m}^{\left( t \right)}\mathbf{F}- \alpha\mathbf{L}_m\mathbf{Z}_{m}^{\left(t \right)}\mathbf{W}+\mathbf{X}_m\mathbf{U}_{m} \right)\right)$\\
  \multirow{1}*{MHSGL-Net} & Feng \textit{et al.} \cite{Feng19Hypergraph}   &   $\min\limits_{\mathbf{Z}_{m}}\sum\limits_{{m=1}}^{M}\frac{1}{2}\Big(\|\mathbf{X}_{m}-\mathbf{Z}_{m}\mathbf{D}_{m}\|_{F}^{2}+\alpha\operatorname{Tr}(\mathbf{Z}_{m}^{T} \mathbf{\tilde{L}}_{m} \mathbf{Z}_{m})\Big)+\beta\|\mathbf{Z}_{m}\|_{2, 1}$ & $\frac{1}{M}\sum\limits_{{m=1}}^{M}\operatorname{\textbf{Prog}}_{\frac{\beta}{L}}\left( \mathbf{Z}_{m}^{\left( t \right)}\left( \mathbf{I}-\frac{1}{L}\mathbf{D}_{m}^{T}\mathbf{D}_m \right) -\frac{\alpha}{\tilde{L}}\mathbf{\tilde{L}}_m\mathbf{Z}_{m}^{\left( t \right)}+\frac{1}{L}\mathbf{X}_m\mathbf{D}_{m}^{T} \right)$ & $\mathcal{T}\left(\mathcal{S}_{\theta}\left(\mathbf{Z}_{m}^{\left( t \right)}\mathbf{F}- \alpha\mathbf{\tilde{L}}_m\mathbf{Z}_{m}^{\left(t \right)}\mathbf{W}+\mathbf{X}_m\mathbf{U}_{m} \right)\right)$\\
\midrule
  \multirow{1}*{Our}  & Above  &  $M_h(\mathbf{x}_{m}):= \sum\limits_{{m=1}}^{M}M_f(\mathbf{x}_{m})+g(\mathbf{z}_{m})+h(\mathbf{z}_{m})$ & Rule (5) & Framework (16) with examples (20)-(19)\\
\bottomrule
\end{tabular}}
\end{table*}

\begin{algorithm}[!htbp]
\caption{A Trustworthy Open-world Learning Protocol (TOLP)}
\label{algorithmTrust}
\begin{algorithmic}[1]
\REQUIRE{Open-world data $\mathbf{X}$ with known labels $\mathbf{Y}$; layer number $T$, training epoch $E$, hyper-parameters $\alpha$, $\lambda_1$ and $\lambda_2$.}
\ENSURE {Perception results $\hat{\mathbf{Y}} \in \{1, \cdots, K; unknown\}$.}
\STATE {Normalize open-world data $\mathbf{X}$ in the range of $[0, 1]$, and initialize parameter set $\mathbf{\Theta}$;}
\STATE {Formalization as Problem (2) for constructing network with design-level interpretability;}
\STATE {Using machine optimization to derive the corresponding Equation (5) with physical meaning;}
\STATE {The preliminary preparations are completed, moving towards the design-level interpretability networks, and implementing the trustworthy protocols;}
\FOR {$e = 1 \rightarrow E$}
  \FOR {$t = 1 \rightarrow T$}
    \STATE {Calculate open-world representation $\mathbf{Z}^{(t)}$ by generalization well-being task-interfaces (6) or (7);}
  \ENDFOR
  \STATE {Compute open-world robustness loss (10);}
  \STATE {Update learnable parameter $\mathbf{\Theta} = \{\mathbf{\Theta}_{\mathbf{z}}, \theta\}$;}
\ENDFOR\\
\STATE {Calculate an agent $a = (a_k+a_u)/2$ and identify by Equation (11) when a new sample appears;}
\RETURN{Perception results $\hat{\mathbf{Y}} \in \{1, \cdots, K; unknown\}$.}
\end{algorithmic}
\end{algorithm}

\begin{algorithm}[!htbp]
\caption{Beyond Single-modal: A Multi-modal Trustworthy Open-world Learning Protocol (MTOLP)}
\label{algorithmTrust2}
\begin{algorithmic}[1]
\REQUIRE{Open-world data $\{\mathbf{X}_{m}\}_{m=1}^{M}$ with known labels $\mathbf{Y}$, layer number $T$, training epoch $E$, hyper-parameters $\alpha$, $\lambda_1$ and $\lambda_2$.}
\ENSURE {Perception results $\hat{\mathbf{Y}} \in \{1, \cdots, K; unknown\}$.}
\STATE {Normalize open-world data $\mathbf{X}_{m}$ in the range of $[0, 1]$, and initialize parameter set $\mathbf{\Theta}$;}
\STATE {Formalization as Problem (2) for constructing $m$-th network with design-level interpretability;}
\STATE {Using machine optimization to derive the corresponding $m$-th Equation (5) with physical meaning;}
\STATE {The preliminary preparations are completed, moving towards the $m$-th design-level interpretability network, and implementing the trustworthy protocols;}
\FOR {$e = 1 \rightarrow E$}
  \FOR {$t = 1 \rightarrow T$}
    \STATE {Calculate the $m$-th open-world representation $\mathbf{Z}_{m}$ by multi-modal generalization well-being task-interfaces (15) or (16);}
     \STATE {Acquire generalized fusion (19) to obtain the open-world co-latent representation $\mathbf{Z}^{(t)}$;}
  \ENDFOR
  \STATE {Compute open-world robustness loss (10);}
  \STATE {Update learnable parameter $\mathbf{\Theta} = \{\mathbf{\Theta}_{\mathbf{z}_{m}}, \mathbf{\Theta}_{\mathcal{F}}, \theta_{m}\}$;}
\ENDFOR\\
\STATE {Calculate an agent $a = (a_k+a_u)/2$ and identify by Equation (11) when a new sample appears;}
\RETURN{Perception results $\hat{\mathbf{Y}} \in \{1, \cdots, K; unknown\}$.}
\end{algorithmic}
\end{algorithm}

\subsection{Theoretical Analysis}\label{sec:the}

\textbf{Proof of Theorem 1.} For any matrix $\mathbf{B} \in \mathbb{R}^{b_1 \times b_2}$, we define the vectorization of the matrix by $vec[\mathbf{B}]$ and the Frobenius norm of the matrix by $\|\mathbf{B}\|_F$.
Let us first state some preliminary explanation for networks in Tables \ref{NetworkExamples}-\ref{NetworkExamples3}.
Convergence can be guaranteed in both single-modal and multi-modal cases (Frameworks (6)-(7) to (15)-(16)), because the latter is a generalization of the former.
On this premise, we define the mapping function $\varphi$ that contains regularizers ($\mathcal{H}_{\theta}$ or $\mathcal{S}_{\theta}$) and fusion $\mathcal{F}$ (if in multi-modal situation).
Then, $\varphi(\mathbf{Z}) = \mathbf{Z}\mathbf{F} - \alpha G(\mathbf{L})\mathbf{Z}\mathbf{W} + H(\mathbf{X}, \mathbf{U})$, which is an abstract universal form.
If SL-Net and SGL-Net can be expressed when $G(\mathbf{L})=\mathbf{L}$, otherwise, MHSL-Net and MHSGL-Net when $G(\mathbf{L})=\mathbf{\tilde{L}}$.
In special cases, it can be degenerate to S-Net and SG-Net when $G(\mathbf{L})=0$.
Then, we want to show that the map $\varphi$ is contraction.
Using the property of the vectorization and the Kronecker product $\otimes$,
\begin{equation}\label{ProblemDerivation}
\begin{aligned}
&\operatorname{vec}[\varphi(\mathbf{Z})]=\operatorname{vec}\left[\mathbf{Z}\mathbf{F}\right] - \alpha\operatorname{vec}\left[G(\mathbf{L})\mathbf{Z}\mathbf{W}\right]+\operatorname{vec}[H(\mathbf{X}, \mathbf{U})]\\&=\left[\left(\mathbf{F}\right)^{\top}\right]\operatorname{vec}[\mathbf{Z}]-\alpha\left[\left(\mathbf{W}\right)^{\top} \otimes G(\mathbf{L})\right] \operatorname{vec}[\mathbf{Z}]+\operatorname{vec}[H(\mathbf{X}, \mathbf{U})]\\&=\alpha\left(\left[\left(\mathbf{F}\right)^{\top}\otimes \mathbf{A}\right]-\left[\left(\mathbf{W}\right)^{\top} \otimes G(\mathbf{L})\right]\right) \operatorname{vec}[\mathbf{Z}]+\operatorname{vec}[H(\mathbf{X}, \mathbf{U})],
\end{aligned}
\end{equation}
where $\mathbf{A} \in \mathbb{R}^{n \times n}$ is a matrix whose diagonal element is $\frac{1}{\alpha}$ and the rest is 0.
Therefore, for any $\mathbf{Z}, \mathbf{Z}^{\prime} \in \mathbb{R}^{n \times c}$,
\begin{equation}\label{ProblemDerivation2}
\begin{aligned}
&\left\|\varphi(\mathbf{Z})-\varphi\left(\mathbf{Z}^{\prime}\right)\right\|_F  =\left\|\operatorname{vec}[\varphi(\mathbf{Z})]-\operatorname{vec}\left[\varphi\left(\mathbf{Z}^{\prime}\right)\right]\right\|_2 =\\&\left\|\alpha\left(\left[\left(\mathbf{F}\right)^{\top}\otimes \mathbf{A}\right] - \left[\left(\mathbf{W}\right)^{\top} \otimes G(\mathbf{L})\right]\right)\left(\operatorname{vec}[\mathbf{Z}]-\operatorname{vec}\left[\mathbf{Z}^{\prime}\right]\right)\right\|_2 \leq \\&\alpha\left\|\left(\left[\left(\mathbf{F}\right)^{\top}\otimes \mathbf{A}\right] - \left[\left(\mathbf{W}\right)^{\top} \otimes G(\mathbf{L})\right]\right)\right\|_2\left\|\operatorname{vec}[\mathbf{Z}]-\operatorname{vec}\left[\mathbf{Z}^{\prime}\right]\right\|_2.
\end{aligned}
\end{equation}
Due to all matrices and layers are normalized, which means the spectral radius of $A = \left[\left(\mathbf{F}\right)^{\top}\otimes \mathbf{A}\right]\in [0, 1]$, $B = \left[\left(\mathbf{W}\right)^{\top} \otimes G(\mathbf{L})\right]\in [0, 1]$, and the range of $\left\|A-B\right\|_2$ is $[0, 1]$,
\begin{equation}\label{ProblemDerivation3}
\begin{aligned}
\left\|\varphi(\mathbf{Z})-\varphi\left(\mathbf{Z}^{\prime}\right)\right\|_F & \leq \alpha\underbrace{\left\|\left(\left[\left(\mathbf{F}\right)^{\top}\otimes \mathbf{A}\right] - \left[\left(\mathbf{W}\right)^{\top} \otimes G(\mathbf{L})\right]\right)\right\|_2}_{[0, 1]}\\&\left\|\operatorname{vec}[\mathbf{Z}]-\operatorname{vec}\left[\mathbf{Z}^{\prime}\right]\right\|_2
\leq \alpha\left\|\mathbf{Z}-\mathbf{Z}^{\prime}\right\|_F.
\end{aligned}
\end{equation}
Since $\alpha \in [0, 1)$, this shows that $\varphi$ is a contraction mapping on the metric space ($\mathbb{R}^{n \times c}, \hat{d}$), where $\hat{d}\left(\mathbf{Z}, \mathbf{Z}^{\prime}\right)=\left\|\mathbf{Z}-\mathbf{Z}^{\prime}\right\|_F$.
Therefore, the convergence of the proposed trustworthy framework is theoretically guaranteed utilizing the Banach fixed-point theorem.

\section{Supplementary of Experiments}\label{Experiments}

In this section, we present some supplementary, including dataset details, compared methods, evaluation indicators, implementation details, parameter settings and other experimental results.

\subsection{Details of Experimental Setup}

\subsubsection{Datasets, Test Settings and Evaluation Metrics}
For single-modal experiments, we employ eight publicly available graph datasets to simulate a single-modal open-world scenario.
Datasets include Chameleon, CoraFull, Cornell, Film, Pubmed, Tesax, UAI and Wisconsin.
Details for all of these eight datasets are presented in Table \ref{singleDataDescription} and below.

\begin{table}[!htbp]
\centering
\caption{A brief description of single-modal graph datasets.}
\resizebox{0.47\textwidth}{!}{
\label{singleDataDescription}
\begin{tabular}{c||c|c|c|c}
\toprule
 Datasets &      \# Nodes & \# Edges  & \# Classes & \# Features \\
 \midrule
Chameleon  & 2,277 & 36,101 & 5 & 2,325\\
CoraFull & 19,793 & 65,311 & 70 & 8,710 \\
Cornell & 183 & 295 & 5 & 1,703 \\
Film & 7,600 & 33,544 & 5 & 932 \\
Pubmed & 19,717 & 44,338 & 3 & 500\\
Tesax & 183 & 309 & 5 & 1,703\\
UAI & 3,067 & 28,311 & 19 & 4,973 \\
Wisconsin & 251 & 499 & 5 & 1,703  \\
\bottomrule
\end{tabular}}
\end{table}

\begin{itemize}

\item \textbf{Chameleon} is Wikipedia networks where nodes represent web pages from Wikipedia and edges indicate mutual links between pages.
    Node feature vectors are bag-of-word representations of informative nouns in the corresponding pages.
    Each node is labeled with one of five classes according to the number of average monthly traffic of the web page.

\item \textbf{CoraFull} is large well-known citation networks where nodes mean documents, and edges are citation links.
    
\item \textbf{Film} is the actor-only induced subgraph of the film-director-actor-writer network. 
    Each nodes correspond to an actor, and the edge between two nodes denotes co-occurrence on the same Wikipedia page.
    Node features correspond to some keywords in the Wikipedia pages.
    We classify the nodes into five categories in term of words of actor's Wikipedia.

\item \textbf{Pubmed} is a citation network of articles related to diabetes from the PubMed database where node attributes are frequency-weighted word frequencies and the labels specify the type of addressed diabetes.
    
\item \textbf{Cornell, Tesax, Wisconsin} come from the WebKB dataset. 
Nodes represent web pages and edges denote hyperlinks between them.
    Node feature vectors are bag-of-word representations of the corresponding web pages.
    Each node is labeled as student, project, course, staff, or faculty.

\item \textbf{UAI} has been utilized in graph convolutional networks for community detection. 
\end{itemize}

For multi-modal experiments, we employ eight publicly available datasets to simulate a multi-modal open-world scenario.
Datasets include Caltech101\footnote{http://www.vision.caltech.edu/Image\_Datasets/Caltech101/}, Hdigit\footnote{https://cs.nyu.edu/$~$roweis/data.html}, MITIndoor\footnote{http://web.mit.edu/torralba/www/indoor.html},
MNIST\footnote{http://yann.lecun.com/exdb/mnist/}, NoisyMNIST\footnote{https://csc.lsu.edu/~saikat/n-mnist/}, 
NUS-WIDE\footnote{https://lms.comp.nus.edu.sg/wp-content/uploads/2019/research/nuswide/NUS-WIDE.html}, Scene15\footnote{https://figshare.com/articles/dataset/15-Scene\_Image\_Dataset/7007177/} and Youtube\footnote{http://archive.ics.uci.edu/ml/machine-learning-databases/00269/}.
Details for all of these eight datasets are presented in Table \ref{multiDatadescription} and below.
\begin{table}[h]
\centering
\caption{A brief description of multi-modal datasets.}
\resizebox{\linewidth}{!}{
\begin{tabular}{c||c|c|c|c}
 \toprule
 Datasets &      \# Samples & \# Modals  & \# Feature Dimensions & \# Classes \\
 \midrule
  Caltech102  & 9,144      & 6    & 48/40/254/1,984/512/928 & 102         \\
  Hdigit & 10,000 & 2 & 784/256 & 10 \\
  MITIndoor        & 5,360    &  4  &  3,600/1,770/1,240/4,096 & 67          \\
  MNIST           & 10,000      & 3    &  30/9/30  & 10         \\
  NoisyMNIST   & 30,000 & 2 & 784/784 & 10 \\
  NUS-WIDE       & 1,600      & 6    & 64/144/73/128/225/500 & 8  \\
  Scene15       & 4,485      & 3    & 1,800/1,180/1,240 & 15  \\
  Youtube & 2,000 & 6 & 2000/1024/64/512/64/647 & 10 \\
  \bottomrule
\end{tabular}}
\label{multiDatadescription}
\end{table}

\begin{itemize}

\item \textbf{Caltech102} is a popular object recognition dataset with 102 classes of images.
Six extracted features are available: 48-D Gabor, 40-D wavelet moments (WM), 254-D CENTRIST,
1,984-D histogram of oriented gradients (HOG), 512-D GIST and 928-D LBP features.

\item \textbf{Hdigit} contains 10,000 images of handwritten numerals for 0 to 9, each image with two sources, i.e., MNIST Handwritten Digits and USPS Handwritten Digits.

\item \textbf{MITIndoor} is a scene dataset which contains 5,360 images with 67 categories. In this dataset, we extract four types of features including 4,096-D PHOW, 3,600-D LBP, 1,770-D CENTRIST, and 1,240-D deep feature.

\item \textbf{NoisyMNIST} is comprised of randomly selected 30,000 samples originating from MNIST image database in ten classes. Therein, the given images come with white Gaussian noise of varied intensities.

\item \textbf{NUS-WIDE} selects eight classes of six feature sets: 64-D color histogram, 225-D block-wise color moments, 144-D color correlogram, 73-D edge direction histogram, 128-D wavelet texture and 500-D bag of words from SIFT descriptors.

\item \textbf{Scene15} contains 15 scene categories with both indoor and outdoor environments, 4,485 images in total.
1,800-D LBP, 1,180-D PHOW, and 1,240-D CENTRIST features are utilized in the experiment.

\item \textbf{Youtube} is a dataset of multi-view video games, samples described by various audio and visual features, including 2,000-D mfcc, 64-D volume stream, 1,024-D spectrogram stream, 512-D cuboids histogram, 64-D hist motion estimate and 647-D HOG features.
\end{itemize}

For each dataset, a portion of the classes are held out as the unknown class and used for testing, while the remaining classes are used as the known classes for training. 
The data is split such that 10$\%$ of the labeled nodes/samples are used for training, 10$\%$ for validation, and 80$\%$ for testing.
%Note that nodes/samples belonging to unknown classes are only included in the testing set. 
The agent for identify the unknown class is determined using the validation set. 
%Similar to traditional semi-supervised nodes/samples classification, the entire graph is fed into the model for each dataset.
The number of unknown classes is varied to evaluate the performance of the models at different proportions of unknown classes.
The accuracy (ACC) is used for evaluation performance.

\subsubsection{Compared Methods}

The compared methods are consititued of feature-based MLP, AE (Autoencoder), ASF-Net \cite{Gui19AFS}, WAST-Net \cite{Sokar22where}, MAE (MultiAE), DUA-Net \cite{Geng21Uncertainty}, TMC-Net \cite{Han2023Trusted}, DSRL-Net \cite{Wang22Learning}; and graph-based GCN \cite{Kipf17Semi}, GAT \cite{Velick18Graph}, SGCN \cite{wu2019simplifying}, FAGCN \cite{Bo21Beyond}, AMGCN \cite{Wang20AMGCN}, HOG-GCN \cite{Wang22Powerful}, GNN-LF \cite{Zhu2021interpreting}, GNN-HF \cite{Zhu2021interpreting}, HGNN$^{+}$ \cite{Gao23HGNN}.
These comparison methods are described as follows.

\textbf{Non-GCN Baseline: MLP, AE, MAE; non-GCN Methods: SF-Net, WAST-Net, DUA-Net, TMC-Net, DSRL-Net.}
\begin{itemize}
\item \textbf{MLP} was a basic forward-structured artificial neural networks, we select it as the baseline.

\item \textbf{AE and MAE} was an automatic learning representation autoencoder network using encoder and decoder, and MAE is a multi-modal version of AE.

\item \textbf{SF-Net} introduced a network-based feature selection architecture, which had an attention module for feature generation and a learning module for the problem modeling. 

\item \textbf{WAST-Net} presented a new efficient unsupervised method for feature selection based on sparse autoencoders. 

\item \textbf{DUA-Net} was guided by the uncertainty of data estimated from the generation view, and intrinsic information from many views was integrated to obtain noise-free representations.

\item \textbf{TMC-Net} provided a paradigm for multi-view learning by dynamically integrating different views at an evidence level.

\item \textbf{DSRL-Net} proposed a block-wise deep neural network with learnable activation functions for learning data-driven sparse regularizers adaptively in an end-to-end manner.

\end{itemize}

\textbf{GCN Baseline: GCN; GCN Variants: SGCN, FAGCN, AMGCN, HOG-GCN.}

\begin{itemize}

\item \textbf{GCN} regularized scalable approach for semi-supervised learning on graph-structured data that was based on a variant of convolutional neural networks.
    
\item \textbf{SGCN} desgined a simplifying GCN through successively removing nonlinearities and collapsing weight matrices between consecutive layers.

\item \textbf{FAGCN} jointly conducted a GCN with the self-gating mechanism, which could adaptively integrate different signals in the process of message passing.

\item \textbf{AMGCN} reformulated the capability of GCNs in fusing topological structures and node features and identify the weakness of GCN.

\item \textbf{HOGGCN} learned the meta-path interactions of different lengths in multiplex heterogeneous networks by multi-layer convolution aggregation.
\end{itemize}

\textbf{GNN Baseline: GAT; GNN Variants: GNN-LF/HF, HGNN$^{+}$.}

\begin{itemize}
\item \textbf{GAT} operated on graph-structured data leveraging masked self-attentional layers to address the shortcomings of prior methods based on graph convolutions.
    
\item \textbf{GNN-LF/GNN-HF} proposed a unified objective optimization framework for graph neural networks with a feature fitting function and a graph term.

\item \textbf{HGNN$^{+}$} constructed a high-order multi-modal/multi-type data correlation modeling graph neural networks to learn an optimal representation.
\end{itemize}

These comparison methods utilize the same settings and losses as the proposed frameworks to test their performance in the open-world, because they have not been tested in an open-world environment.
As for the multi-modal comparison methods, we extend the above algorithms to multi-modal scenarios, where each modality utilizes the corresponding networks to learn a sub-modal latent representation, and then uses the four fusion methods employed in this paper to select the optimal one as the most comparative open-world perceived results.
Training losses for semi-supervised classification are (10) in the main body.

\subsubsection{Implementation Details}

All proposed open-world frameworks are implemented with Pytorch on a standard Ubuntu-16.04 operation system with a NVIDIA Tesla P100 GPU.
%We keep other parameters consistent to ensure the fairness of the experiments.
Two single-modal and four multi-modal proposed networks include SL-Net, SGL-Net, MSL-Net, MSGL-Net, MHSL-Net, and MHSGL-Net are compared with other competitors as shown in Tables 1-3, and the training losses are (10) in the main body of paper.
The number of training epochs is set as 200, the learning rate is set to 0.001, and the number of layers is set the optimal such as tested in Figures 5-8.
The results in Table 5 consider the performance with the best fusion effect.
The multi-modal adjacency graphs $\mathbf{S}_m$ are constructed by $k$-NN methods.
The construction of Laplacian and hyper-Laplacian matrices $\mathbf{L}_m$ and $\mathbf{\tilde{L}}_m$ are followed by \cite{Xie18Hyper}.
%Other method-specific parameters are the same as the suggestions in the original paper.

\subsection{Experimental Supplementary of Tables and Figures and Other Results}\label{sec:dat}

Tables \ref{SingleAblationStudyNetworkTable}-\ref{MultiAblationStudyNetworkTable} and Figures \ref{TSNEVisualizationMSLNet1}-\ref{lossclassification} are supplementary to the main body of original paper.

\begin{table*}[t]
\centering
\caption{Trustworthiness study of interpretability of the trustworthy frameworks on single-modal tasks.}
\resizebox{\linewidth}{!}{
\begin{tabular}{c|c||cccccccc}
\toprule \multicolumn{2}{c||}{Datasets} & Chameleon & CoraFull & Cornell & Film & Pubmed & Tesax & UAI & Wisconsin\\
\midrule
\multirow{2}*{Unknown = 1} & S-Net & 37.74 & 32.78 & 69.13 & 28.17 & 60.34 & 69.13 & 55.28 & 76.24 \\
                    & SL-Net & 37.14 & \textbf{\color{red}44.95} & 64.43 & \textbf{\color{red}28.49} & \textbf{\color{red}61.54} & 66.44 & 51.83 & 70.30 \\
\midrule
\multirow{2}*{Unknown = 2} & S-Net & 53.28 & 36.37 & 82.28 & 42.64 & 76.19 & 82.28 & 51.69 & 74.76\\
                    & SL-Net & \textbf{\color{red}53.76} & \textbf{\color{red}42.85} & 80.38 & \textbf{\color{red}52.51} & 63.39 & 81.01 & 48.79 & 67.48 \\
\midrule
\multirow{2}*{Unknown = 1} & SG-Net & 36.44 & 34.58 & 61.07 & 28.14 & 59.23 & 61.07 & 53.53 & 72.77 \\
                    & SGL-Net & \textbf{\color{red}38.98} & \textbf{\color{red}44.79} & \textbf{\color{red}63.09} & \textbf{\color{red}28.55} & \textbf{\color{red}61.45} & \textbf{\color{red}63.09} & \textbf{\color{red}55.44} & \textbf{\color{red}73.76} \\
\midrule
\multirow{2}*{Unknown = 2} & SG-Net & 53.13 & 30.90 & 86.08 & 40.35 & 42.65 & 86.08 & 49.56 & 74.76 \\
                    & SGL-Net & \textbf{\color{red}54.02} & \textbf{\color{red}44.32} & \textbf{\color{red}86.71} & \textbf{\color{red}52.51} & 40.07 & 82.91 & \textbf{\color{red}51.41} & 73.30\\
\bottomrule
\end{tabular}}
\label{SingleAblationStudyNetworkTable}
\end{table*}

\begin{table*}[t]
\centering
\caption{Trustworthiness study of interpretability of the trustworthy frameworks on multi-modal tasks.}
\resizebox{\linewidth}{!}{
\begin{tabular}{c|c||cccccccc}
\toprule \multicolumn{2}{c||}{Datasets} & Caltech102 & Hdigit & MITIndoor & MNIST & NoisyMNIST & NUS-WIDE & Scene15 & Youtube\\
\midrule
\multirow{3}*{Unknown = 1} & MS-Net & 47.76 & 91.90 & 38.63 & 80.69 & 74.71 & 39.78 & 59.31 & 49.54 \\
                        & MSL-Net & \textbf{\color{red}48.39} & \textbf{\color{red}93.79} & \textbf{\color{red}40.26} & \textbf{\color{red}90.21} & \textbf{\color{red}77.89} & 38.40 & 59.09 & \textbf{\color{red}51.82} \\
                        & MHSL-Net & 42.86 & 92.62 & 40.70 & 85.37 & 74.38 & \textbf{\color{red}40.48} & \textbf{\color{red}60.69} & 50.83 \\
\midrule
\multirow{3}*{Unknown = 3} & MS-Net & 46.99 & 90.07 & 36.90 & 87.18 & 80.32 & 47.94 & 64.06 & 56.68 \\
                    & MSL-Net & \textbf{\color{red}48.44} & \textbf{\color{red}96.56} & 39.94 & \textbf{\color{red}91.85} & \textbf{\color{red}84.72} & 48.39 & \textbf{\color{red}64.47} & \textbf{\color{red}61.40} \\
                    & MHSL-Net & 39.31 & 90.90 & \textbf{\color{red}42.10} & 88.59 & 73.22 & \textbf{\color{red}48.39} & 63.92 & 58.02 \\
\midrule
\multirow{3}*{Unknown = 1} & MSG-Net & 44.36 & 86.55 & 39.23 & 80.86 & 67.89 & 36.55 & 56.65 & 47.50 \\
                        & MSGL-Net & 45.56 & \textbf{\color{red}91.98} & \textbf{\color{red}42.54} & \textbf{\color{red}89.74} & \textbf{\color{red}71.51} & 37.09 & 55.99 & \textbf{\color{red}47.79} \\
                        & MHSGL-Net & \textbf{\color{red}49.06} & 90.81 & 39.53 & 88.88 & 67.78 & \textbf{\color{red}38.09} & \textbf{\color{red}58.20} & 46.88 \\
\midrule
\multirow{3}*{Unknown = 3} & MSG-Net & 42.84 & 91.00 & 36.11 & 87.56 & 81.47 & 46.36 & 62.39 & 56.99 \\
                    & MSGL-Net & 43.62 & \textbf{\color{red}95.02} & \textbf{\color{red}40.26} & \textbf{\color{red}91.83} & \textbf{\color{red}83.60} & \textbf{\color{red}47.56} & 62.31 & \textbf{\color{red}61.74} \\
                    & MHSGL-Net & \textbf{\color{red}47.84} & 92.58 & 37.60 & 91.40 & 75.23 & 46.81 & \textbf{\color{red}62.64} & 56.87 \\
\bottomrule
\end{tabular}}
\label{MultiAblationStudyNetworkTable}
\end{table*}

\setcounter{figure}{0}
\begin{figure*}[!htbp]
  \centering
  \includegraphics[width=\linewidth]{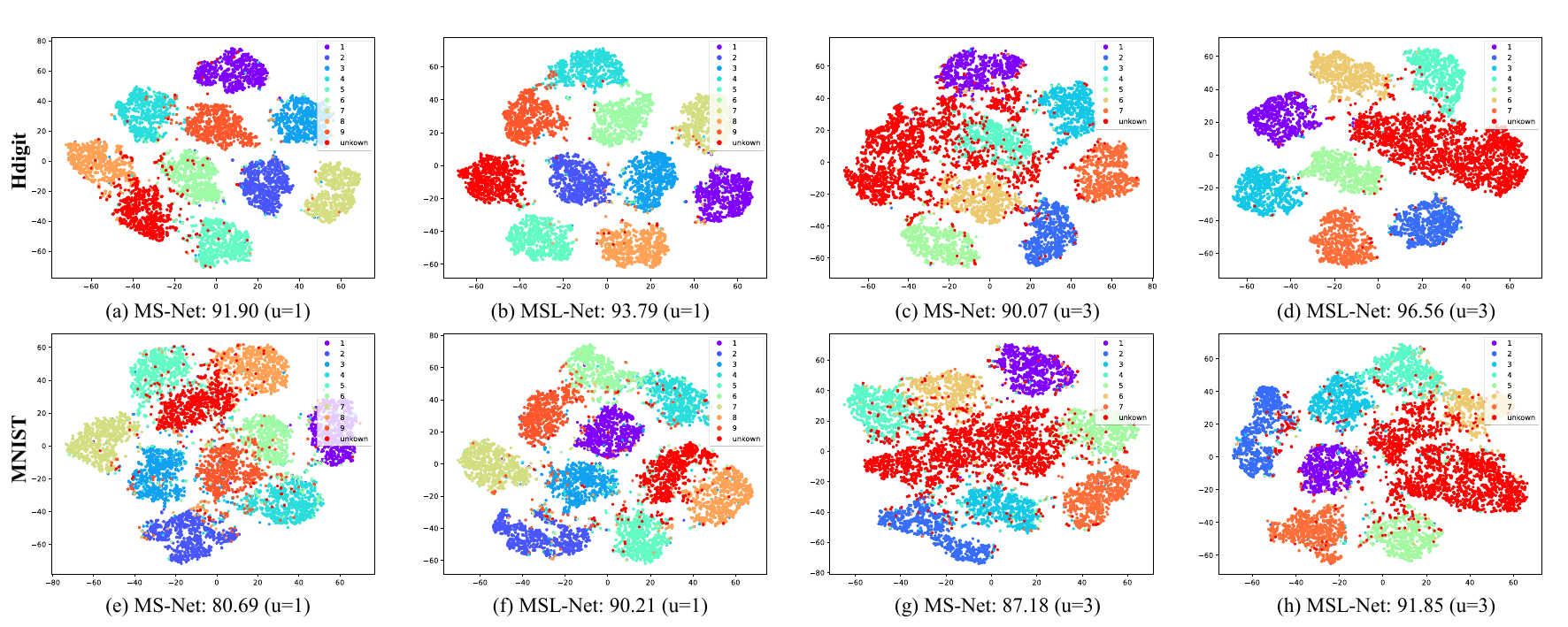}\\
  \caption{A t-SNE visualization of MSL-Net of Hdigit and MNIST on the multi-modal semi-supervised classification (Evaluation: accuracy; ``u" is the number of unknown classes).}
  \label{TSNEVisualizationMSLNet1}
\end{figure*}

\begin{figure*}[!htbp]
  \centering
  \includegraphics[width=\linewidth]{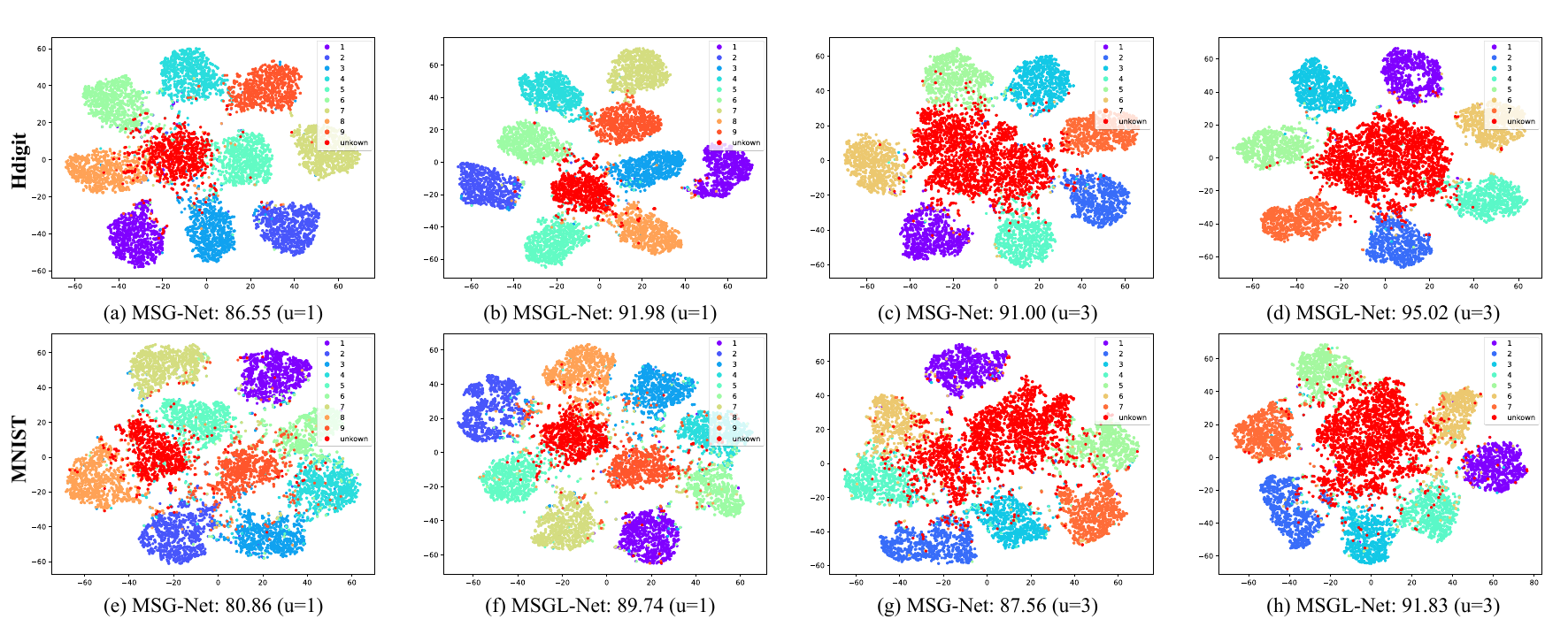}\\
  \caption{A t-SNE visualization of MSGL-Net of Hdigit and MNIST on the multi-modal semi-supervised classification (Evaluation: accuracy; ``u" is the number of unknown classes).}
  \label{TSNEVisualizationMSGLNet}
\end{figure*}

\begin{figure*}[t]
  \centering
  \includegraphics[width=\linewidth]{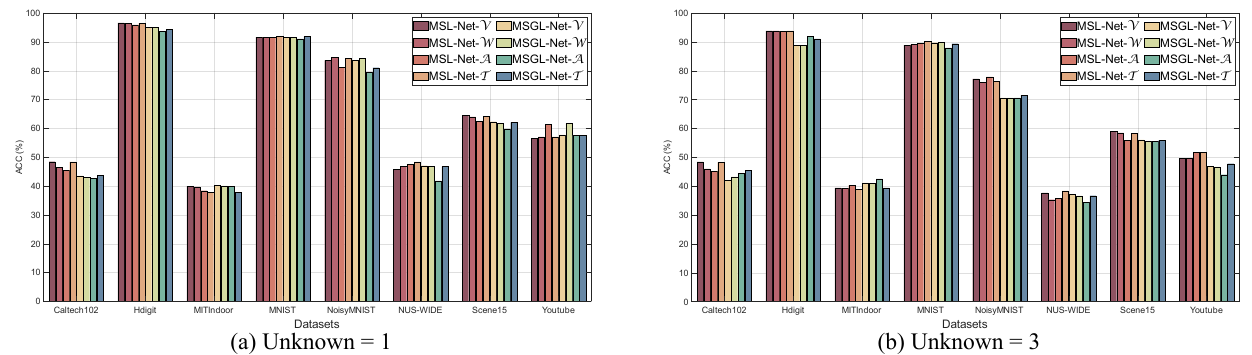}\\
  \caption{Ablation study of generalized fusion ways of trustworthy frameworks (MSL-Net and MSGL-Net).}
  \label{paramultifusion1}
\end{figure*}

\begin{figure*}[t]
  \centering
  \includegraphics[width=\linewidth]{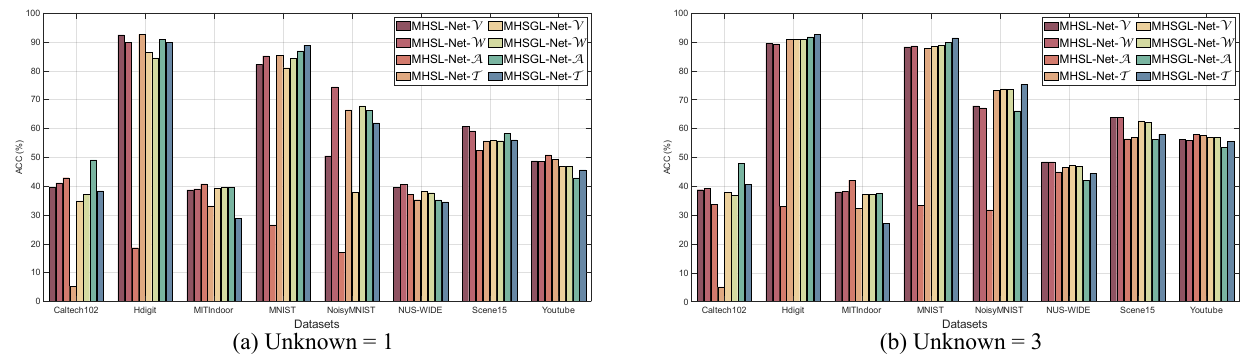}\\
  \caption{Ablation study of various fusion ways of trustworthy frameworks (MHSL-Net and MHSGL-Net).}
  \label{paramultifusion2}
\end{figure*}

\begin{figure*}[t]
  \centering
  \includegraphics[width=\linewidth]{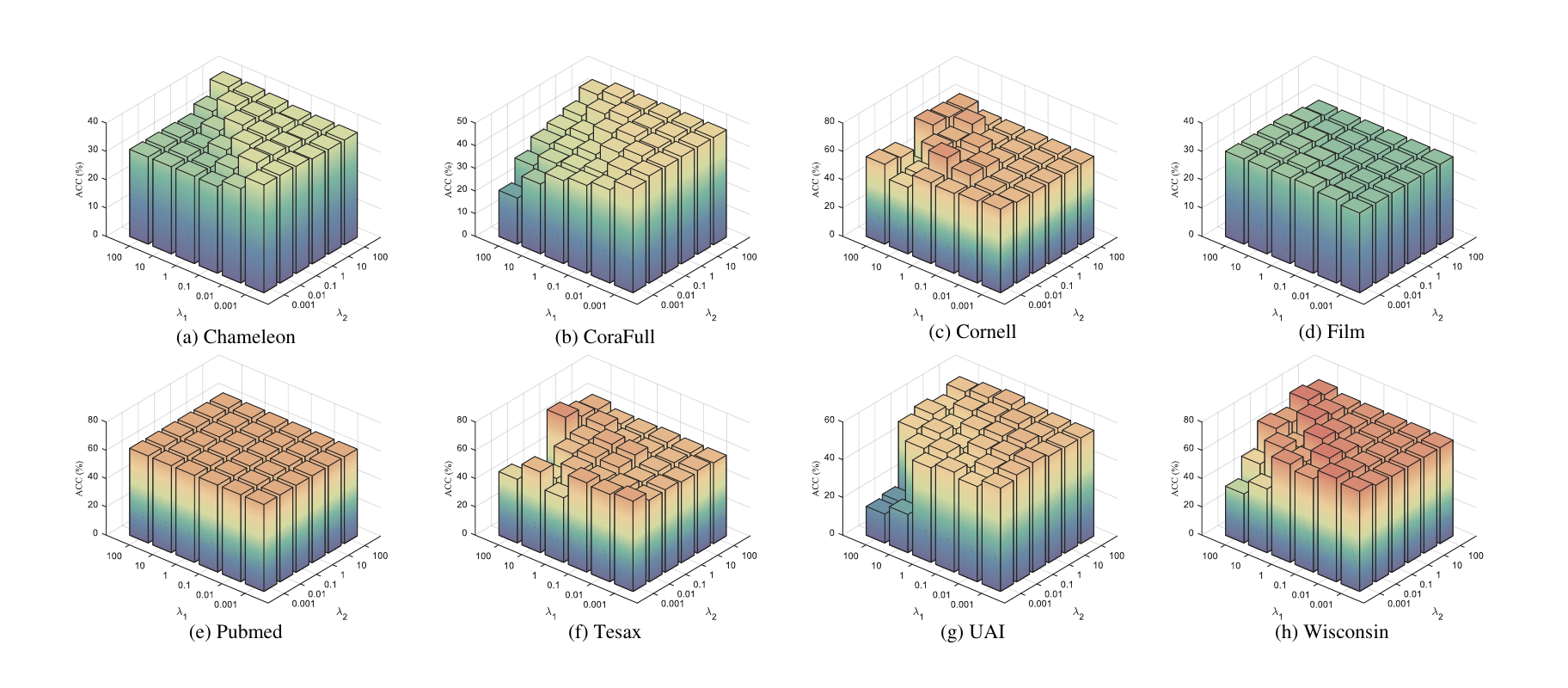}\\
  \caption{Robustness analysis of hyper-parameters $\lambda_1$ and $\lambda_2$ of SL-Net in ranges of $[0.001, 0.01, 0.1, 1, 10, 100]$ on single-modal semi-supervised classification tasks.}
  \label{parasingle1}
\end{figure*}

\begin{figure*}[t]
  \centering
  \includegraphics[width=\linewidth]{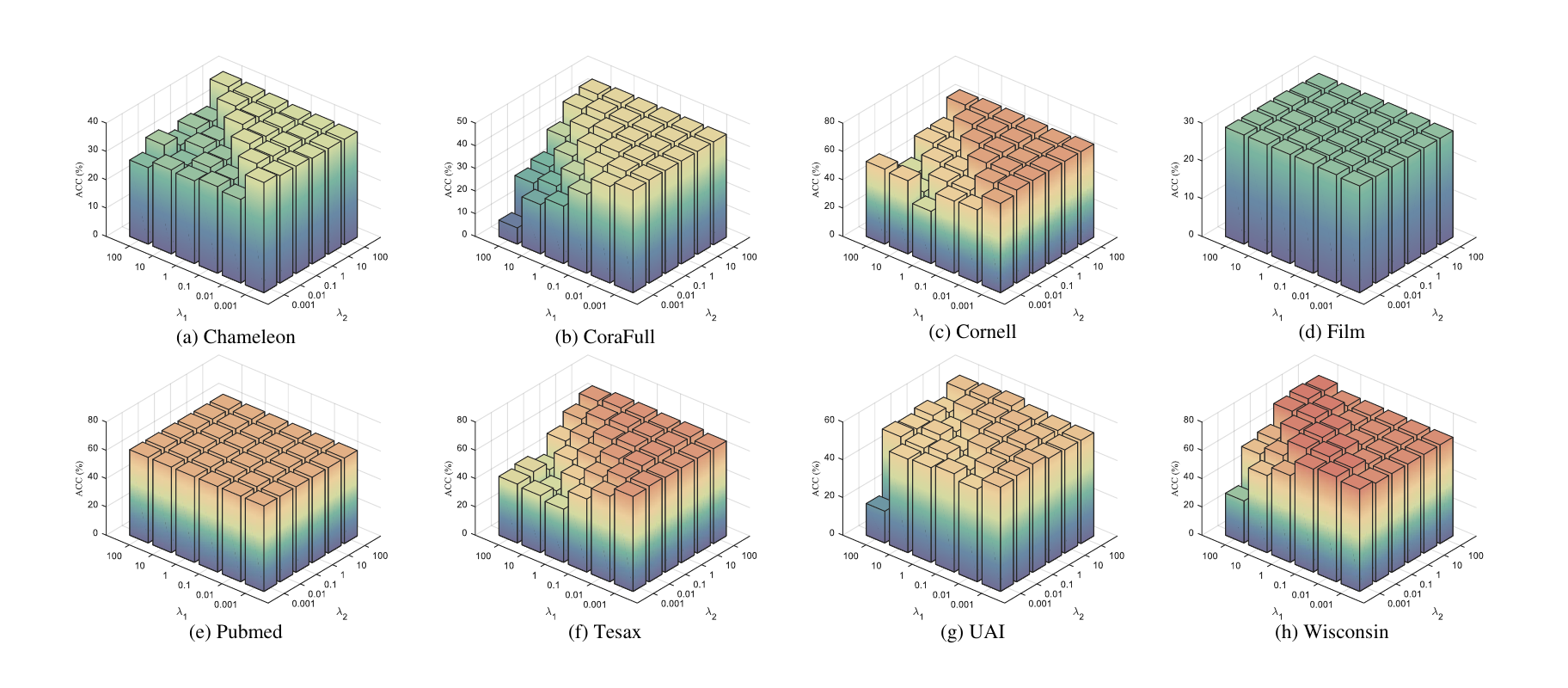}\\
  \caption{Robustness analysis of hyper-parameters $\lambda_1$ and $\lambda_2$ of SGL-Net in ranges of $[0.001, 0.01, 0.1, 1, 10, 100]$ on single-modal semi-supervised node classification tasks.}
  \label{parasingle2}
\end{figure*}

\begin{figure*}[t]
  \centering
  \includegraphics[width=\linewidth]{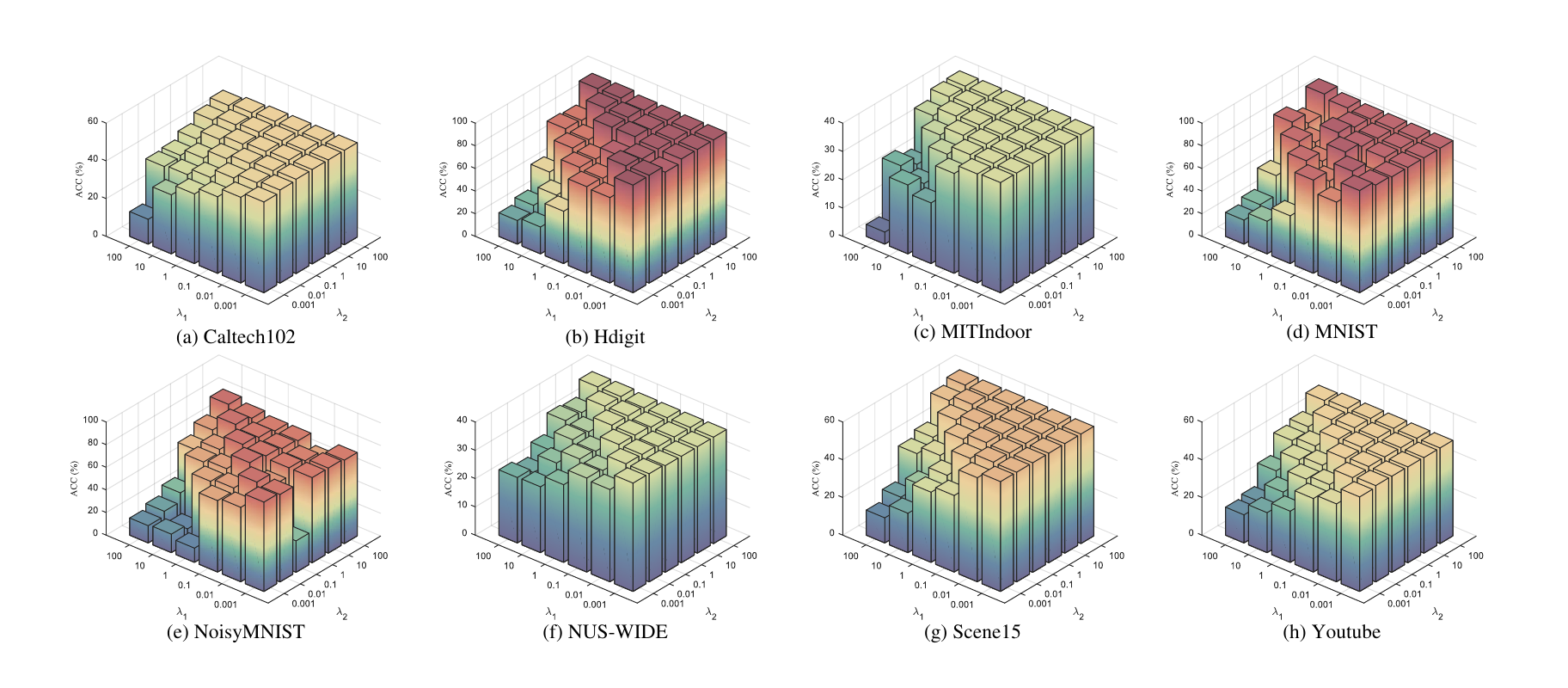}\\
  \caption{Robustness analysis of hyper-parameters $\lambda_1$ and $\lambda_2$ of MSL-Net in ranges of $[0.001, 0.01, 0.1, 1, 10, 100]$ on multi-modal semi-supervised classification tasks.}
  \label{paramulti1}
\end{figure*}

\begin{figure*}[t]
  \centering
  \includegraphics[width=\linewidth]{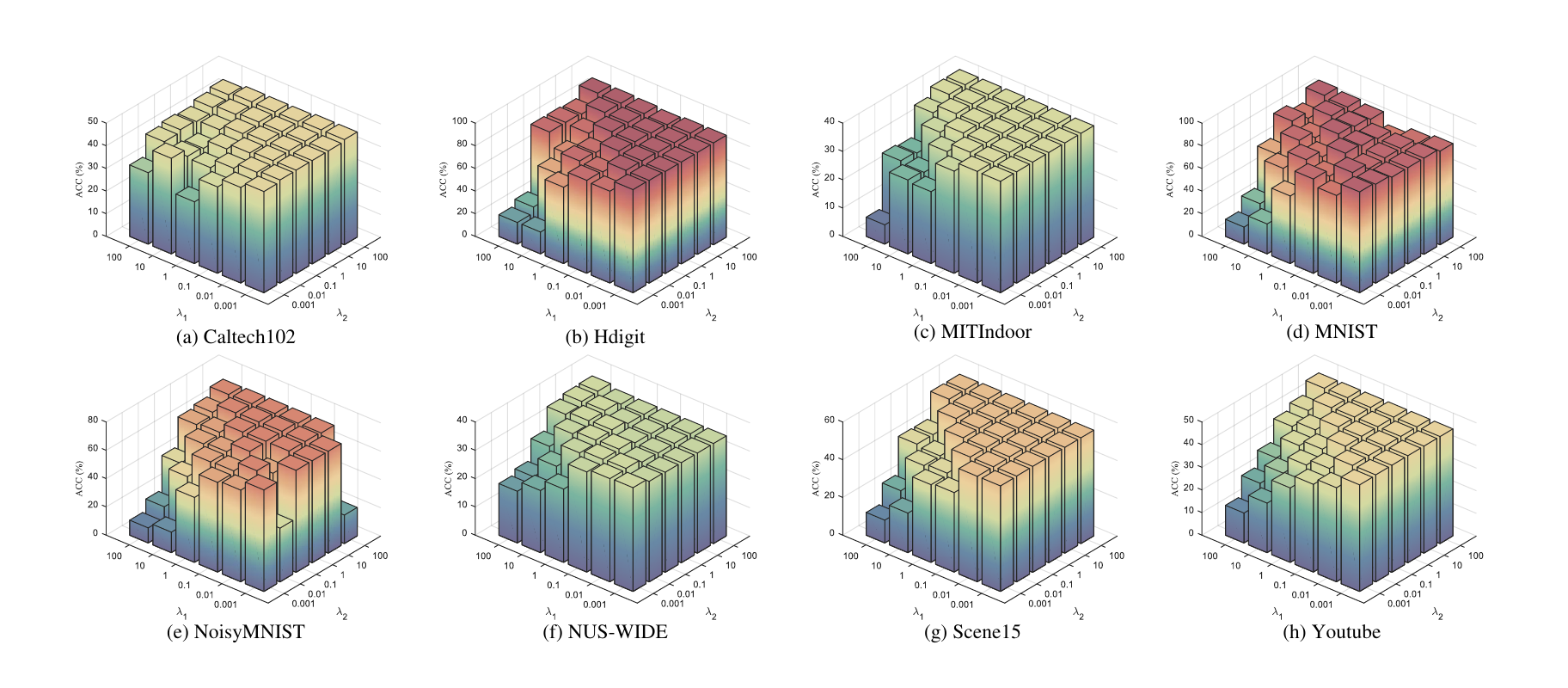}\\
  \caption{Robustness analysis of hyper-parameters $\lambda_1$ and $\lambda_2$ of MSGL-Net in ranges of $[0.001, 0.01, 0.1, 1, 10, 100]$ on multi-modal semi-supervised classification tasks.}
  \label{paramulti2}
\end{figure*}

\begin{figure*}[t]
  \centering
  \includegraphics[width=\linewidth]{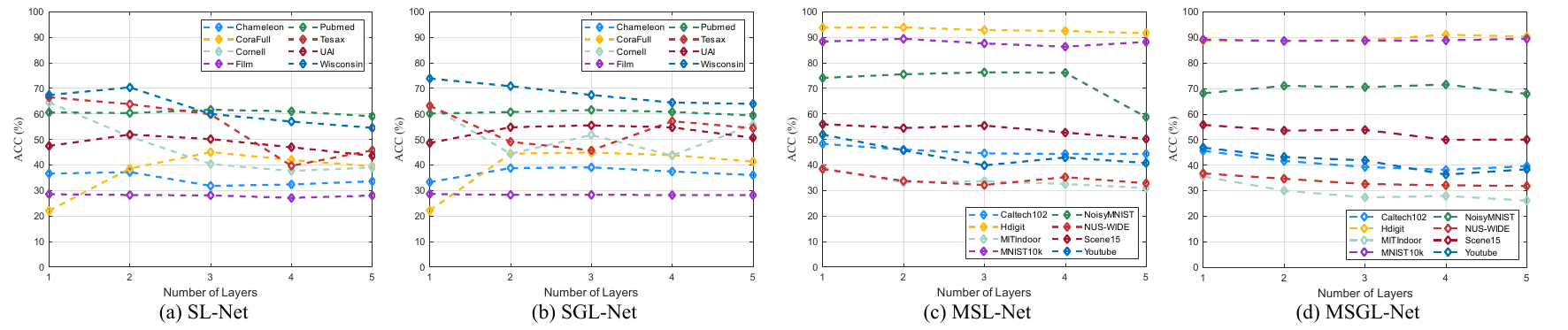}\\
  \caption{Parameter sensitivity analysis of networks derived from the proposed trustworthy frameworks in terms of layer impact on single-modal and multi-modal semi-supervised classification tasks.}
  \label{blockclassification}
\end{figure*}

\begin{figure*}[t]
  \centering
  \includegraphics[width=\linewidth]{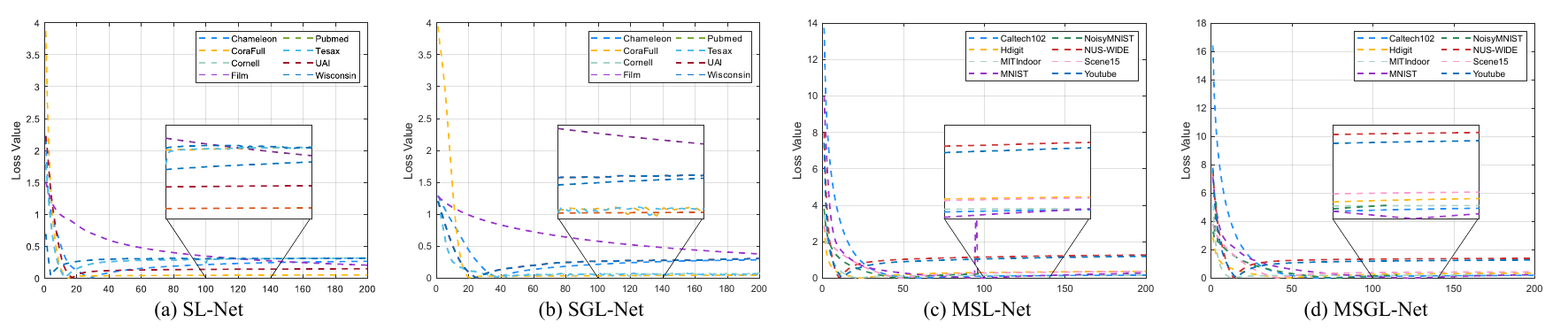}\\
  \caption{Parameter sensitivity analysis of networks derived from the proposed trustworthy frameworks in terms of convergence behavior on single-modal and multi-modal semi-supervised classification tasks.}
  \label{lossclassification}
\end{figure*}

%%%%%%%%%%%%%%%%%%%%%%%%%%%%%%%%%%%%%%%%%%%%%%%%%%%%%%%%%%%%%%%%%%%%%%%%%%%%%%%%%%%%%%%%%%%%%%%%

\bibliographystyle{ACM-Reference-Format}
\bibliography{ML}
\end{document}